\documentclass[10pt,journal,compsoc]{IEEEtran}

\ifCLASSOPTIONcompsoc
  \usepackage[nocompress]{cite}
\else
  \usepackage{cite}
\fi
\usepackage[T1]{fontenc}
\usepackage{graphicx}
\usepackage{amsmath}
\usepackage{url}
\usepackage{amssymb}
\usepackage{booktabs}
\usepackage{tikz}
\usepackage{etoolbox}
\usepackage{ragged2e}
\usepackage{enumitem}
\usepackage{caption}
\usepackage{multirow}
\newcommand*{\circled}[1]{\lower.7ex\hbox{\tikz\draw (0pt, 0pt)%
    circle (.5em) node {\makebox[1em][c]{\small #1}};}}
\robustify{\circled}
\usepackage[colorlinks,linkcolor=blue]{hyperref}

\newcommand{\ie}{\textit{i.e.}}
\newcommand{\eg}{\textit{e.g.}}
\newcommand{\et}{\textit{et al.}}
\newcommand{\ting}[1]{\textcolor{black}{#1}}

\ifCLASSINFOpdf
\else
\fi
\begin{document}
%
\title{DaGAN++: Depth-Aware Generative Adversarial Network for Talking Head Video Generation}
%
%
%
%

\author{Fa-Ting~Hong,~Li Shen,~and~Dan~Xu~\IEEEmembership{Member,~IEEE}
\IEEEcompsocitemizethanks{\IEEEcompsocthanksitem Fa-Ting Hong and Dan Xu are with the Department
of Computer Science and Engineering, The Hong Kong University of Science and Technology, Hong Kong SAR.
\protect
E-mail: \{fhongac, danxu\}@cse.ust.hk
\IEEEcompsocthanksitem Li Shen is with Alibaba Group. E-mail: lshen.lsh@gmail.com.
}
}

%
%

\markboth{Journal of \LaTeX\ Class Files,~Vol.~14, No.~8, August~2015}%
{Shell \MakeLowercase{\textit{et al.}}: Bare Demo of IEEEtran.cls for Computer Society Journals}


\IEEEtitleabstractindextext{%
\begin{abstract}
\justifying
Predominant techniques on talking head generation largely depend on 2D information, including facial appearances and motions from input face images. Nevertheless, dense 3D facial geometry, such as pixel-wise depth, plays a critical role in constructing accurate 3D facial structures and suppressing complex background noises for generation. However, dense 3D annotations for facial videos is prohibitively costly to obtain. In this work, firstly, we present a novel self-supervised method for learning dense 3D facial geometry (\ie, depth) from face videos, without requiring camera parameters and 3D geometry annotations in training. We further propose a strategy to learn pixel-level uncertainties to perceive more reliable rigid-motion pixels for geometry learning. Secondly, we design an effective geometry-guided facial keypoint estimation module, providing accurate keypoints for generating motion fields. Lastly, we develop a 3D-aware cross-modal (\ie, appearance and depth) attention mechanism, which can be applied to each generation layer, to capture facial geometries in a coarse-to-fine manner. Extensive experiments are conducted on three challenging benchmarks (\ie, VoxCeleb1, VoxCeleb2, and HDTF). The results demonstrate that our proposed framework can generate highly realistic-looking reenacted talking videos, with new state-of-the-art performances established on these benchmarks. The codes and trained models are publicly available on the \href{https://github.com/harlanhong/CVPR2022-DaGAN}{GitHub project page}.

\end{abstract}

}

\maketitle

\IEEEdisplaynontitleabstractindextext

%
\IEEEpeerreviewmaketitle

\IEEEraisesectionheading{\section{Introduction}\label{sec:introduction}}
\IEEEPARstart{T}{alking} head video generation, which targets generating a realistic talking head video via given one still source image and one dynamic driving video, has attracted rapidly increasing attention from the communities in recent years. 
A preferred talking head generation model should enable the still source image to accurately imitate rich facial expressions and complex head movements presented from the target driving video. 
This naturally triggers many practical applications, 
\eg~AI-based human conversation, digital human broadcast, and virtual anchors in films.

\begin{figure}[t]
  \centering
    \includegraphics[width=1\columnwidth]{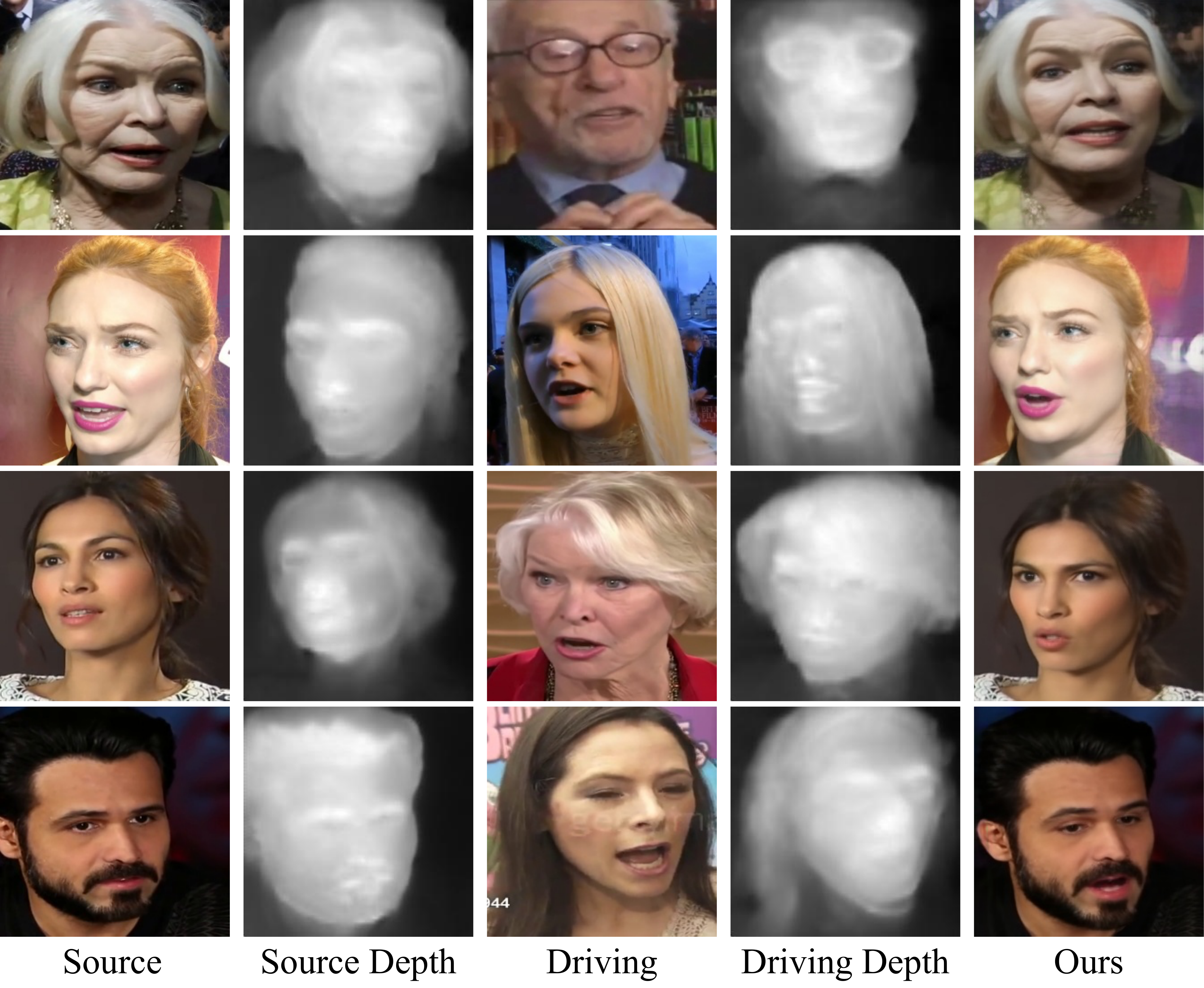}
\caption{Our proposed face depth learning network can produce accurate face depth maps. By incorporating accurate face depth maps, our DaGAN++ can produce high-quality generations facilitated by the depth maps.
    }
    \label{fig:firstpage}       
\end{figure}

Existing methods for talking head generation can be generally divided into two categories: model-based and model-free models. Model-based methods~\cite{yao2020mesh, yin2022styleheat, zakharov2019few, zhao2021sparse, ren2021pirenderer, zhang2022metaportrait} typically utilize pretrained facial models (\eg~3DMM~\cite{tran2018nonlinear} or landmark detectors~\cite{wood20223d}) to obtain disentangled parameters of the human face or the landmarks which contain both facial shape and expression information. In this way, they can obtain appearance-irrelevant motion information using these disentangled parameters or landmarks. However, these kinds of methods can inevitably suffer from error accumulation caused by the inaccuracy of the pretrained models. This problem is even more serious when the training data distribution of the facial 3D model is inconsistent with that of the talking head model. Therefore, model-free methods~\cite{siarohin2019first, wang2021one, siarohin2021motion} do not rely on pre-trained third-party models, and extend their methods for arbitrary objects, even for cartoon objects. For instance, FOMM~\cite{siarohin2019first} proposes a first-order approximation of motion using self-supervised detected keypoints to estimate motion flows between two faces. Based on the first-order method, Face-vid2vid~\cite{wang2021one} extends motion estimation to 3d feature space to enable head movement controllably.
Although promising performances are achieved, these existing works mainly focus on learning effective facial appearance and motion representations to perform the talking head generation. However, 3D \emph{dense} facial geometry is a critical clue for face generation since it essentially captures rich 3D facial structures and details.

Intuitively, incorporating facial geometry information can provide clear benefits for talking-head video generation. First of all, given the fact that the videos capture head movements within 3D physical environments, the 3D geometry can substantially contribute to the accurate reconstruction of 3D facial structures, and thus the capability to preserve realistic 3D facial structures is important for generating high-fidelity facial videos. Secondly, dense geometry can facilitate the model in distinguishing noisy background information during generation, particularly in the presence of cluttered backgrounds. Lastly, dense geometry proves particularly useful for identifying subtle, expression-related facial movements. However, a significant challenge in employing 3D dense geometry to enhance the generation is that, the 3D geometry annotations for this specific task are generally unavailable, while 3D labeling of video data is prohibitively costly.

\par Based on the above-mentioned motivations, we propose learning pixel-wise depth maps using geometric warping and photometric consistency in a self-supervised manner. Our proposal allows for the automatic reconstruction of dense 3D facial geometry by training with consecutive face frames in videos, eliminating the need for expensive 3D geometry annotations. More specifically, we develop a new face depth estimation network to produce accurate facial depth maps (see Fig.~\ref{fig:firstpage}). 
In talking head generation, the moving face contains both rigid and non-rigid motions. The non-rigid motion is usually related to expression-related movements, and most of the face pixels are associated with rigid motions.
To learn more accurate rigid motion, 
we introduce a scheme to learn an uncertainty map to {discover} reliable {rigid-motion pixels} that indicate more significant and important rigid motions for face depth estimation. The uncertainty map is then constrained on the photometric consistency loss for geometry learning.

Based on the learned dense facial depth maps, we further design two effective strategies to utilize face geometry for enhancing talking face synthesis.
The first strategy is geometry-guided facial keypoint detection.
As the facial keypoints are utilized to compute facial motion for pose transfer, an accurate estimation of the keypoints is thus critically important.
Since the face depth map explicitly represents the 3D facial structure information, we combine geometry representations from depth maps and appearance representations from face images together, to predict more accurate and structure-consistent facial keypoints.
The second strategy is a designed cross-modal attention mechanism, which aims to boost the learning of the face motion field by suppressing the noisy motion from the background and enhancing expression-related micro-movements. The proposed cross-modal attention is a geometry-guided attention to impose dense 3D geometric constraints on the motion field. The cross-modal attention module is also applied at each stage of the generation process to achieve a course-to-fine geometry guided generation.


This paper significantly extends our previous CVPR version, \ie~DaGAN~\cite{hong2022depth}. Specifically, we propose a new enhanced facial depth estimation network, devise a more robust geometry-enhanced multi-layer generation process, further elaborate on related works, provide additional technical details of the different components, and largely expand the experiments and analysis to verify the effectiveness on several talking head datasets, including an additional new high-resolution talking-head dataset HDTF~\cite{zhang2021flow}. We coin the new and performance-enhanced framework as DaGAN++.
Compared to DaGAN, the improved face depth network with uncertainty as guidance for geometry learning can generate more accurate face depth maps, which helps capture and reproduce small changes of facial expressions, leading to more expressive and realistic talking head animations. By replacing the vanilla generation module in DaGAN, the geometry-enhanced multi-layer generation with course-to-fine geometric guidance can also advance facial generation details. Based on all the new designs, DaGAN++ achieves clearly improved results upon DaGAN, and establishes new state-of-the-art results on different challenging benchmarks.
\par To summarize, the contribution of this paper is threefold:
\begin{itemize}
    \item To the best of our knowledge, our work is the first to introduce a self-supervised geometry learning approach for recovering dense 3D facial geometry information (\ie, facial depth maps) from face videos, for talking head video generation without the requirement of camera parameters or expensive 3D annotations. We show the learned face depth maps can effectively enhance face video generation.

    \item We propose a novel talking head generation framework DaGANN++, which integrates learned dense face depths into the generation network through two designed strategies: the geometry-guided facial keypoint estimation to capture the accurate motion of the human face, and the geometry-enhanced multi-layer generation that benefits from facial geometry guided message passing among features in a course-to-fine manner.

    \item We conduct extensive experiments on three challenging datasets, demonstrating that the proposed model significantly surpasses previous methods by incorporating more precise geometry information and exhibits superior generation performance across all datasets compared to state-of-the-art techniques.

\end{itemize}

{We organize the rest of the paper as follows}. Section~\ref{sec:related} presents a review of closely related literature. A comprehensive illustration of the DaGAN++ framework for talk head video generation is introduced in Section~\ref{sec:methodology}. Section~\ref{sec:experiments} shows the experimental results and their interpretation. {The conclusion of this paper is made in Section~\ref{sec:conclusion}.}


\section{Related Works}\label{sec:related}
\subsection{Talking Head Synthesis}

We can divide the talking head generation methods into two categories based on driven modalities: image-driven and audio-driven talking head generation. 

\noindent\textbf{Audio-driven talking head generation.} Audio-driven talking head generation methods~\cite{zhou2019talking, wang2021audio2head, deng2020disentangled, zhou2021pose} utilize audio inputs to establish connections between the audio space and the expression or lip movement space. 
Recent developments in end-to-end cross audio-visual generation~\cite{chen2017deep,zhao2018sound,gao20192,zhao2019sound,xu2019recursive,gan2020music,gan2020foley} led to an exploration of speaker-independent settings~\cite{chung2017you,song2018talking,kr2019towards,zhou2019talking,zhou2020makelttalk}, aiming for a universal model that handles multiple identities using minimal frame references. DAVS~\cite{zhou2019talking} proposed an innovative method using disentangled audio-visual representations to capture subject-related and speech-related information, generating realistic sequences and improving lip reading for audio-video retrieval tasks. However, as pose information is challenging to infer from audio, and individuals can utter the same words in different poses, PC-AVS~\cite{zhou2021pose} introduced a video sequence to provide pose information for the generated video, allowing expression control while maintaining consistent lip motion. {Besides that, SPACE~\cite{gururani2023space} maps audio to controllable facial landmarks and generates videos using a pre-trained face generator to achieve expression control. Similarly, MODA~\cite{liu2023moda} controls the expression through dense facial landmarks, which are detected by a pre-trained Mediapipe model~\cite{lugaresi2019mediapipe}.}
Nevertheless, some methods~\cite{yao2022dfa,liu2022semantic,guo2021ad} collaborate with the neural radiance field (NeRF~\cite{mildenhall2021nerf}) for the audio-driving talking head to create a novel view of talking head.
AD-NeRF~\cite{guo2021ad} synthesizes both the head and upper body regions, producing natural results while allowing adjustments in audio signals, viewing directions, and background images. 
SSP-NeRF~\cite{liu2022semantic}  introduces a novel representation for imitable human avatars, combining a morphable model with feed-forward networks to improve reconstruction quality and novel-view synthesis.  
Additionally, several methods~\cite{kim2022diffface, shen2023difftalk} utilize the diffusion model~\cite{ho2020denoising} to produce synthetic image. Diffused Head~\cite{stypulkowski2023diffused} proposes an autoregressive diffusion model for talking face generation, which takes an image and an audio sequence as input and produces realistic head movements, facial expressions, and background preservation. 
DiffTalk~\cite{shen2023difftalk} employs reference facial images and landmarks to facilitate personality-conscious general synthesis, adeptly producing high-resolution, audio-driven talking head videos for previously unseen identities, eliminating the need for fine-tuning. 

Our method falls into the video-driven talking head generation category, as it utilizes video frames as the driving target to obtain motion and expression information. In contrast, audio-driven talking head generation relies on audio input as the primary information source for generating facial expressions and lip movements. \ting{While audio-driven methods use audio signals as a driving force, our DaGAN++ uses images or videos as the driving mechanism. Due to this fundamental difference in the type of input, we are unable to conduct a direct comparison between our method and audio-driven methods. Normally, the audio-driven talking head methods can primarily influence the lip shape based on the input audio; however, it cannot precisely model facial expressions solely using audio due to ambiguities. Thus, video-driven methods serve as crucial complements in achieving comprehensive facial control.}

\noindent\textbf{Video-driven talking head generation.} Video-driven talking head generation methods~\cite{wang2021one, siarohin2019first, wiles2018x2face, zhang2019one, burkov2020neural, yao2020mesh,hong2022depth} take the videos as the driven modality to extract the motion information. Several video-driven talking head generation methods~\cite{yao2020mesh,yin2022styleheat,ren2021pirenderer,wang2021safa} employ 3D morphable model (\eg~3DMM~\cite{blanz1999morphable}) to extract an expression code for motion modelling. Specifically, StyleHeat~\cite{yin2022styleheat} and PIRenderer~\cite{ren2021pirenderer} utilize the 3DMM model~\cite{booth20163d,deng2019accurate} to generate an expression code for the driving face and then feed it into the motion field learning module. Based on the 3DMM model, they can animate the driving expression, and also control the pose of the generated face by modifying the rotation or translation matrix. Additionally, some other video-driven methods~\cite{zhang2022metaportrait,zhao2021sparse} consider using pre-trained landmark detectors to detect facial landmarks to represent the expression of the face. In~\cite{zhao2021sparse}, it presents an efficient and effective method for generating realistic face animations from a single image and sparse landmarks, by unifying global and local motion estimation to faithfully transfer motion, while {enhancing landmark detection in videos} for temporally coherent and high-quality results. MetaPortrait~\cite{zhang2022metaportrait} introduces an ID-preserving talking head generation framework that leverages dense landmarks for accurate geometry-aware flow fields, and adaptively fuses source identity during synthesis for better preservation of key characteristics. {HyperReenact~\cite{bounareli2023hyperreenact} utilizes a 3D shape model to encode the facial poses of faces, and these facial poses are then utilized to estimate the parameter residual for StyleGAN2.}  
Besides these third-party model-based methods, some video-driven talking head generation methods~\cite{siarohin2019first,hong2022depth,wang2021one,siarohin2021motion,zhao2022thin} attempt to learn keypoints of the human face to represent the facial expression in a self-supervised manner. FOMM~\cite{siarohin2019first} introduces a self-supervised image animation framework {that decouples appearance and motion information and computes the motion between two faces by using their keypoints.} The subsequent keypoint-based works~\cite{wang2021one,siarohin2019first} adopt FOMM's motion modelling method. For instance, Face-vid2vid~\cite{wang2021one} extends 2D keypoints into 3D to control the head pose by modifying the rotation and translation elements. {
MCNet~\cite{hong2023implicit} learns a memory bank to compensate for the warped features, producing images with large poses. Additionally, some methods~\cite{gong2023toontalker, wang2022latent} learn an implicit motion space to achieve motion transfer. LIA~\cite{wang2022latent} decomposes facial motion into a combination of base motions. Based on these base motions, Toontalker~\cite{gong2023toontalker} utilizes a transformer~\cite{liu2021swin} to align motions from different domains into a common latent space.}

Image-driven talking head generation remains an active research area with various methods and techniques that have been proposed to tackle this problem. A significant difference of our proposed method is to learn and  incorporate dense facial depth  estimation in a self-supervised geometry learning manner. We develop an effective geometry-aware talking head generation framework while not requiring any camera parameters or expensive 3D annotations, and our approach clearly advances the generation results compared to the discussed related works. 
\begin{figure*}[t]
  \centering
    \includegraphics[width=1\textwidth]{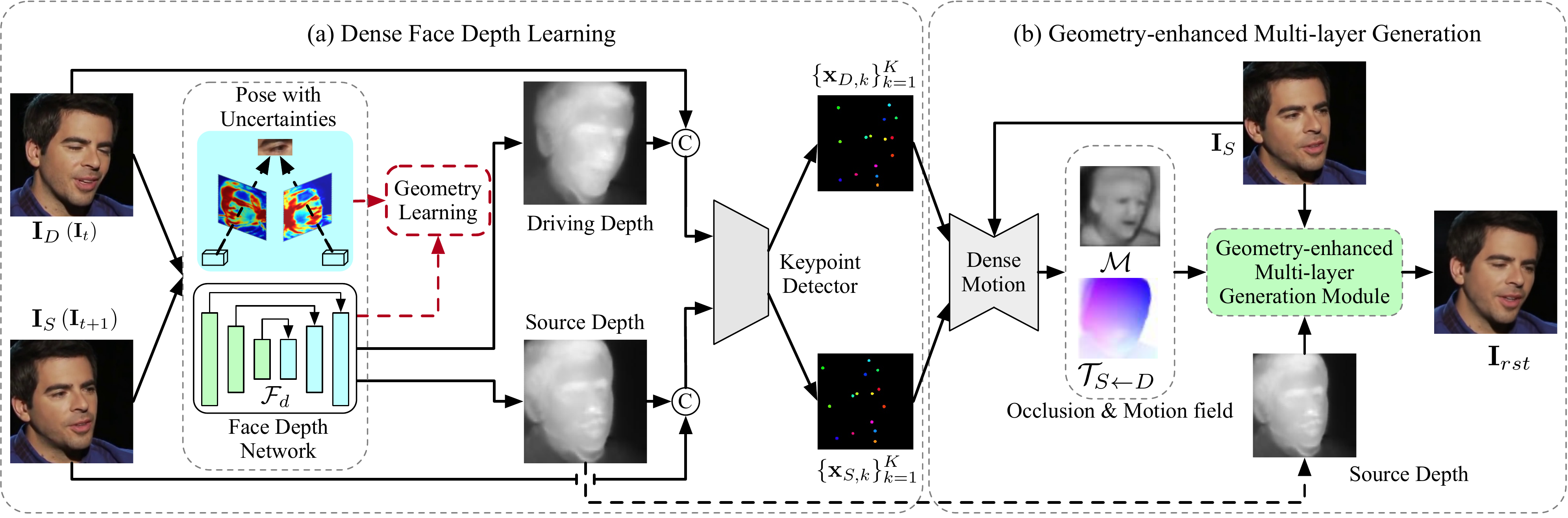}
    \caption{The framework overview of our DaGAN++ approach. We can mainly divide the process of DaGAN++ into three components: (a) An uncertainty-aware face depth learning network. We learn a facial depth network in a self-supervised manner to recover dense 3D facial geometry utilizing only face videos without the need for camera parameters and 3D annotations. (b) Geometry-guided facial keypoint detection. We utilize the face depth network to estimate the depth maps of the face images and feed them with the corresponding RGB images into the keypoint detector to estimate face keypoints.
    (c) A geometry-enhanced multi-layer generation process. After learning the motion field and an occlusion map between two faces by the detected keypoints, we embed the facial geometry into each layer in the generation process through cross-modal geometry-guided attention.
    }
    \vspace{-10pt}
    \label{fig:framework}       
\end{figure*}

\subsection{Self-supervised depth estimation} 
The estimation of depth in monocular images or videos has recently gathered wide interest from both industry and academia. Several investigations have been carried out in the literature to address challenges of depth prediction along the direction~\cite{fu2018deep,xu2018unsupervised, digging2019monodepth2,ha2016high,luo2020consistent,xu2017multi,zhou2017unsupervised,gordon2019depth,Madelving}. For instance, Zhou~\et~\cite{zhou2017unsupervised} utilized an end-to-end learnable method with view synthesis as the supervisory signal to predict depth maps from monocular video sequences without using 3D supervision. Built upon~\cite{zhou2017unsupervised}, Clement~\et~\cite{digging2019monodepth2} achieved notable improvements by incorporating a minimum projection loss that alleviates occlusion issues between frames, and an auto-masking loss to ignore ambiguous stationary pixels. Gordon~\et~\cite{gordon2019depth} sought to learn camera intrinsic parameters for every pair of successive frames, allowing the model to function when the camera parameters are unknown in new environments.

\par Instead of dealing with natural scenes, our work introduces, for the first time, a self-supervised face depth estimation method for face reenactment and synthesis in videos, purely based on video images without requiring any camera parameters or 3D annotations. The learned depth maps provide rich 3D facial geometry, which is utilized in our model for accurate keypoint detection and for guiding the talking face generation in a coarse-to-fine process. 

\section{Methodology}\label{sec:methodology}
In this paper, we utilize a source image and a driving video as input to generate a synthetic video that maintains the identity information from a provided source image while mimicking the facial movements of an individual in a driving video. We first design a robust facial depth estimation technique based on a self-supervised learning framework, using only training face videos without relying on any costly 3D geometry annotations. By incorporating accurate geometric information, our proposed DaGAN++ can effectively integrate facial geometry into the generation process, yielding higher-quality videos that capture better facial structures and detailed expression-related micro-movements.
\vspace{-5pt}
\subsection{Overview}
As shown in Fig.~\ref{fig:framework}, given an input source image $\mathbf{I}_S$ and a driving video $\mathbf{V}_D$ that contains a sequence of frames $\{\mathbf{I}_D^1, \mathbf{I}_D^2, \dots, \mathbf{I}_D^T\}$, the proposed deep generation framework DaGAN++
aims to produce a sequence of synthetic images $\{\mathbf{I}_{rst}^1, \mathbf{I}_{rst}^2, \dots, \mathbf{I}_{rst}^T\}$. Our proposed DaGAN++ primarily comprises three major components: (i) A dense face depth estimation network $\mathcal{F}_d$. Taking two consecutive frames from a face video as two different views, we learn depth estimation {by employing a self-supervised learning approach. Then our} DaGAN++ is jointly trained, keeping $\mathcal{F}_d$ fixed. 
(ii) Geoemtry-guided keypoints-based motion estimation. Given a source image $\mathbf{I}_S$ and a driving target image $\mathbf{I}_D^t$ (omitting the superscript $t$ in subsequent notations for simplicity), we employ the learned face depth network $\mathcal{F}_d$ to generate depth maps ($\mathbf{D}_s$ and $\mathbf{D}_d$) for both the source and target image. Next, we take both the face geometry (depth maps) and appearance information (RGB images) into consideration to detect face keypoints (\ie~$\{\mathbf{x}_{S,k}\}^K_{k=1}$ and $\{\mathbf{x}_{D,k}\}^K_{k=1}$); (iii) We then utilize the detected keypoints to compute motion fields between two face images using the Taylor approximation approach~\cite{siarohin2019first}. Then, the geometry-enhanced multi-layer generation process takes in the estimated motion field, the depth map $\mathbf{D}_s$ of the source image, and the $N$ encoded feature maps $\{\mathbf{F}^i_e\}_{i=1}^N$, which are extracted by a CNN encoder with the source image $\mathbf{I}_S$ as input, and output enhanced feature maps for generating the final output image. In each layer, to encourage the model to maintain facial structure details, we additionally learn dense cross-modal attention using the source depth map $\mathbf{D}_s$ and the warped source feature map $\mathbf{F}_w^i$ that is the output of $\mathbf{F}_e^i$ through warping by the motion flow $\mathcal{T}_{S \leftarrow D}$. In this manner, we can incorporate geometry information into the generation process, to yield high-quality structure-preserved output.
\begin{figure*}[t]
  \centering
    \includegraphics[width=1\linewidth]{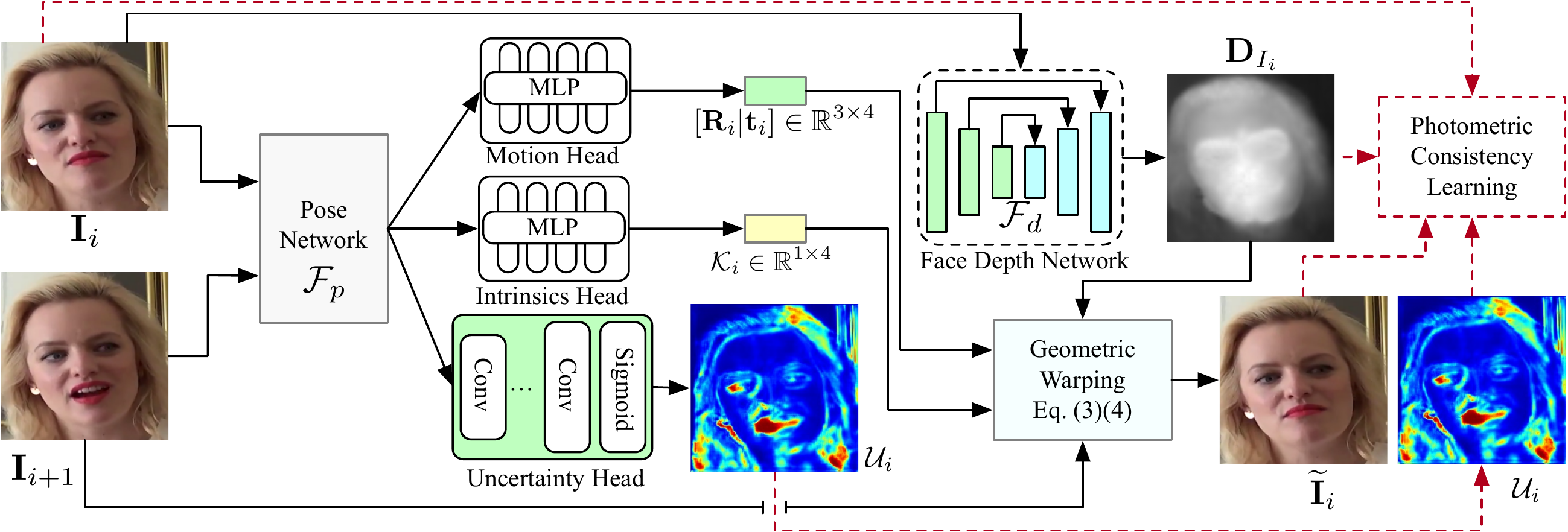}
    \caption{The pipeline of our face depth learning network. We design three different decoding heads after the pose backbone network. They estimate the relative camera poses {$[\mathbf{R}_i|\mathbf{t}_i]$}, the camera intrinsic matrix $\mathcal{K}_i$, and an uncertainty map $\mathcal{U}_i$, respectively. Finally, we learn the face depth network by optimizing a photometric consistency loss which is constructed via the reference image $\mathbf{I}_i$ and the reconstructed image $\mathbf{\widetilde{I}}_{i}$ that is obtained from a geometry warping module.} 
   
    \label{fig:depth_estimate}  
\end{figure*}
\vspace{-5pt}
\subsection{Dense face depth learning}
In this section, we introduce our proposed face depth estimation network, which is capable of automatically recovering depth information of human faces. While the SfM-Learner~\cite{zhou2017unsupervised} previously proposed an unsupervised method for learning scene depth, it is conducted in outdoor scenarios in an autonomous driving context. However, it remains unclear how to effectively learn face depth estimation from videos in which the data distributions are significantly different from the outdoor scenarios, and the cameras are typically static. Besides, SfM-Learner still requires provided camera intrinsic parameters in learning. In our talking head generation task, the face videos are probably directly from the internet, the camera intrinsics are thus not available. Therefore, we develop an approach to learn dense facial depths based on ordinary face videos for talking head generation, without requiring any camera parameters and 3D annotations.

\subsubsection{Self-supervised face depth estimation} 
As can be observed in Fig.~\ref{fig:depth_estimate}, there are two main modules in our facial depth learning network, namely, a face depth sub-network $\mathcal{F}_d$ and a pose sub-network $\mathcal{F}_p$. The face depth network is optimized using available training face videos. Specifically, we first extract two consecutive video frames $\mathbf{I}_i$ and $\mathbf{I}_{i+1}$, and take the former as a target view while the latter as a source view. Our facial depth learning network aims to predict several geometric elements, which include a depth map $\mathbf{D}_{I_i}$ for the target image $\mathbf{I}_{i}$, the camera intrinsic matrix $\mathcal{K}_{I_i\rightarrow I_{i+1}}$ for each input image pair, a relative camera pose $\mathbf{R}_{I_i\rightarrow I_{i+1}}$ with translation $\mathbf{t}_{I_i\rightarrow I_{i+1}}$ between the two images, and an uncertainty map $\mathcal{U}_{I_i\rightarrow I_{i+1}}$ that indicates the probability of movement occurrence. 
In our method, $\mathcal{K}_{I_i\rightarrow I_{i+1}}$ is learned in a pair-specific manner, so the input only requires video frames. For simplicity, {we replace the subscript ``$I_i\rightarrow I_{i+1}$'' with ``$i$'' in our subsequent notations}.

\par The depth map $\mathbf{D}_{I_i}$ can be produced by the face depth network $\mathcal{F}_d(\cdot)$. The motion matrix (including the rotation $\mathbf{R}_i$ and the translation $\mathbf{t}_i$), the camera intrinsic matrix $\mathcal{K}_i$, and the uncertainty map $\mathcal{U}_i$ are predicted from the three different heads respectively in the pose network $\mathcal{F}_p(\cdot)$:
\begin{gather} \label{eq:estimate_geo}
    \mathbf{D}_{I_i} =  \mathcal{F}_d(\mathbf{I}_i), \\
    [\mathbf{R}_i, \mathbf{t}_i],
    \mathcal{K}_i, \mathcal{U}_i = \mathcal{F}_p(\mathbf{I}_i \ || \ \mathbf{I}_{i+1}), 
\end{gather}
where the symbol $||$ represents a concatenation operation along the channel dimension.
Consequently, utilizing the geometry elements estimated above, we perform a photometric projection to obtain a reconstructed source view by warping the target view as follows:
\begin{gather}
    \mathbf{q}_\mu \sim
    \mathcal{K}_i [\mathbf{R}_i \, | \,\mathbf{t}_i] \mathbf{D}_{I_i}(\mathbf{p}_j)\mathcal{K}_i^{-1}\mathbf{p}_j  \\
    \mathbf{\widetilde{I}}_{i} = \mathcal{B_I}(\mathbf{I}_{i+1}, \{\mathbf{q}_{\mu}\}_{\mu=1}^M),
\end{gather}
where $\mathbf{q}_{\mu}$ and $\mathbf{p}_{j}$ represent the warped pixel of the source image $\mathbf{I}_{i+1}$ and the original pixel of the target image $\mathbf{I}_{i}$, respectively; $M$ denotes the number of pixels in the image; $\mathcal{B_I}(\cdot)$ is a differentiable bilinear interpolation function; and $\mathbf{\widetilde{I}}_{i}$ is the reconstructed image. Consequently, we can establish a photometric consistency error $\mathcal{L}_{Pe}(\cdot,\cdot)$ between $\mathbf{\widetilde{I}}_{i}$ and $\mathbf{I}_{i}$, which enables training the face depth network without additional 3D annotations for supervision.

\begin{figure*}[t]
  \centering
    \includegraphics[width=1\linewidth]{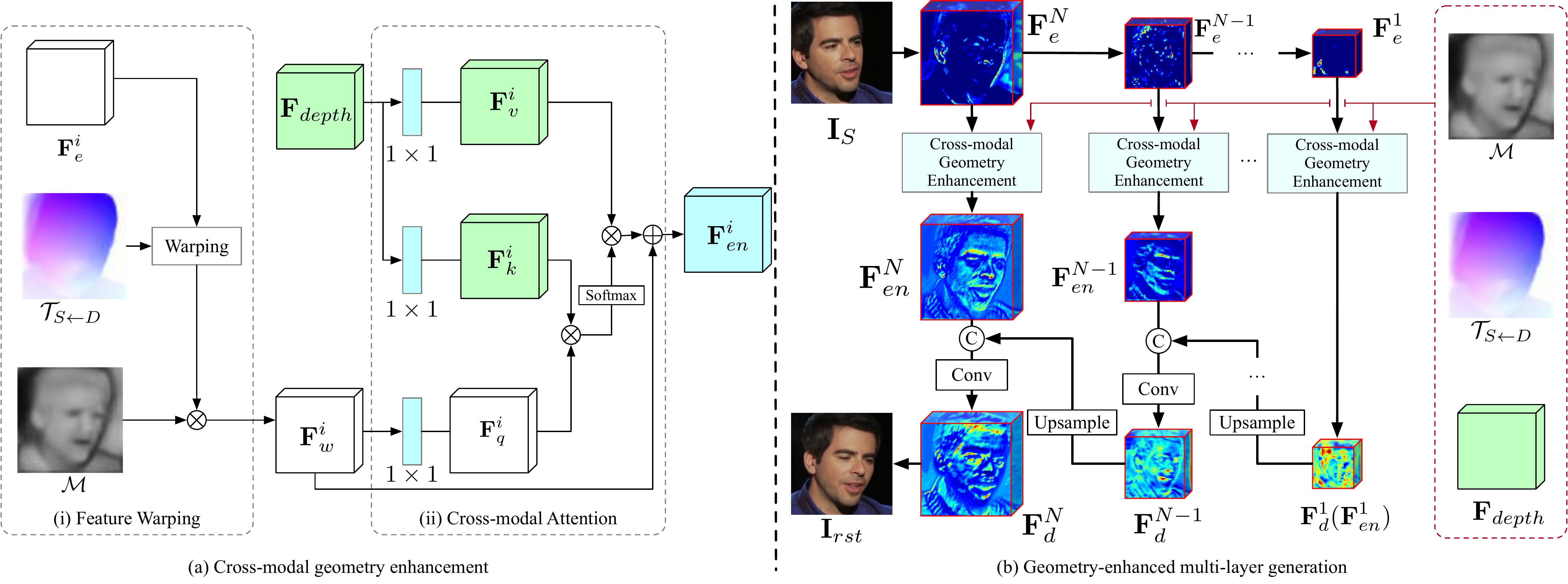}
    \caption{Illustration of our geometry-enhanced multi-layer generation process. We propose a cross-modal geometry enhancement mechanism to learn geometry-enhanced generation face features. In the cross-modal geometry enhancement module (see a), we first transfer the pose of the driving face to the source face by warping encoded feature $\mathbf{F}_e^i$ with the motion flow $\mathcal{T}_{S\leftarrow D}$. Then we learn an occlusion mask $\mathcal{M}$ to represent the region possibly occluded because of large motion. We finally apply the cross-modal attention mechanism to produce geometry-enhanced feature $\mathbf{F}_{en}^i$ at each layer for a coarse-to-fine generation (see b). 
    The symbols $\bigotimes$, $\bigoplus$, and \circled{c} represent the matrix multiplication, and pixel-wise matrix addition, and concatenation operation along the channel dimension, respectively.}
    \label{fig:depth_attention}  
\end{figure*}
\subsubsection{Uncertainty-guided face depth estimation} 
In talking head video generation, the driving target videos are typically captured with static cameras, 
and thus the backgrounds of the videos remain motionless.
{The moving pixels of the face in the image can also be divided into rigid-motion and non-rigid-motion pixels. Generally, the rigid-motion pixels dominate the whole face, and present more reliable and effective motion for geometry learning.}
To better learn the rigid motion {between two views}, we propose a mechanism by estimating a pixel-wise uncertainty map that indicates probabilities of reliable movements occurring at each pixel position. Then, we employ the uncertainty map in the final objective function to regularize the pixels with a high probability of movement occurring. More specifically, following~\cite{digging2019monodepth2}, we first construct a photometric consistency error $\mathcal{L}_{Pe}$ as follows:
\begin{equation} \label{eq:pe_loss}
\begin{aligned}
\mathcal{L}_{Pe}(\mathbf{I}_{i}, \mathbf{\widetilde{I}}_{i}) =& \alpha (1-SSIM(\mathbf{I}_{i}, \mathbf{\widetilde{I}}_{i})) \\
&+ (1-\alpha)||\mathbf{I}_{i} - \mathbf{\widetilde{I}}_{i}||,
\end{aligned}
\end{equation}
{where the $SSIM(\cdot,\cdot)$~\cite{wang2004image} measures an apperance similary of two images.} In this work, $\alpha$ is set as 0.8 to achieve the best results.
To produce a smooth depth map, 
we also employ a smoothness constraint \cite{gordon2019depth} on the depth learning by
\begin{equation}
    \mathcal{L}_{D} = \sum_{x,y}|\partial_xD_{(x,y)}|e^{-|\partial_xI_{(x,y)}|}+|\partial_yD_{(x,y)}|e^{-|\partial_yI_{(x,y)}|}.
\end{equation}
As discussed before, to utilize the dominated and more reliable rigid pixel motions for geometry learning, we consider the estimated pixel-level uncertainty map $\mathcal{U}_i$ as guidance to filter out uncertain {non-rigid} pixel regions:
\begin{equation}\label{eq:depth_loss}
    \mathcal{L}_{depth} = \lambda_1\frac{{\mathcal{M}_{a}}}{\mathcal{U}_i}\mathcal{L}_{Pe}(\mathbf{I}_{i}, \mathbf{\widetilde{I}}_{i})+\lambda_2\mathcal{L}_{D}+\lambda_3\log{\mathcal{U}_i},
\end{equation}
where the hyper-parameters $\lambda_1$, $\lambda_2$ and $\lambda_3$ control the balance of each optimization objective. {Since the camera is static in our face videos, the background always remains static while only the face shows motion. Therefore, an auto-masking $\mathcal{M}_{a}$ \cite{digging2019monodepth2} is introduced to ignore those stationary pixels in the image, and enable our geometry learning model to focus on the motion region.} The last term $\log{\mathcal{U}}_i$ is a regularization for the uncertainty map. In this work, we adopt a stack of upblocks to upsample the output of the pose network, followed by a sigmoid function to produce a soft uncertainty map (see Fig.~\ref{fig:structure}f):
\begin{equation}
    \mathcal{U}_i = \mathrm{Sigmoid}(\mathcal{F}_{\mathcal{U}}(\mathcal{F}_p(\mathbf{I}_i, \mathbf{I}_{i+1})))
\end{equation}
where $\mathcal{F}_{\mathcal{U}}(\cdot)$ is the ucertainty head and $\mathcal{F}_p(\cdot,\cdot)$ represents the pose network in Fig.~\ref{fig:depth_estimate}. The ``$\mathrm{Sigmoid}(\cdot)$'' denotes the sigmoid function. \ting{It is noteworthy that the talking head generation task focuses on the foreground and aims to generate high-quality and realistic face video with a good alignment of driving motion. Therefore, the background of the talking head dataset is typically considered static under the basic setup of talking head generation.  Importantly, it should be acknowledged that the performance of the face depth network could be affected by the presence of dynamic background objects or noise across multiple frames. Estimating reliable geometric elements (\ie, relative camera pose $R_i$ and translation $t_i$ in Eq.~\ref{eq:estimate_geo}) becomes problematic if both the background and foreground become dynamic during training. Such a situation would conflict with the assumptions of SFM as outlined in \cite{zhou2017unsupervised}, potentially leading to inaccurate results. However, this issue is mitigated in our work because, generally, the background of the training data remains stable in this specific task.}

After training the face depth network with the overall loss function $\mathcal{L}_{depth}$, we employ $\mathcal{F}_d$ to produce dense facial depth maps for source and target images, which are further utilized by DaGAN++.
\ting{\subsubsection{Motion modeling by geometry-guided keypoints}}
\label{sec:feature warping} 
The keypoints of the face are important for estimating motion fields from a target driving video. To generate more accurate motion fields, 
we propose to utilize learned face depth maps to provide geometry information for learning the keypoints.
Once the face depth map is obtained from the face depth network, {the keypoint detector accepts a combined input consisting of the face RGB image and its associated depth map}, and predicts a set of sparse keypoints ($\mathbf{x}_{S,k},\mathbf{x}_{D,k} \in \mathbb{R}^{1\times 2}$) and their corresponding Jacobian metrics ($J_{S,k},J_{D,k} \in \mathbb{R}^{2\times 2}$) of the human face (see Fig.~\ref{fig:framework}b).
\begin{equation}
    \{\mathbf{x}_{\tau,k}, J_{\tau,k}\}_{k=1}^K = \mathcal{F}_{kp}(\mathbf{I}_\tau||\mathbf{D}_\tau), \tau\in\{S, D\},
\end{equation}
where subscript $\tau$ indicates whether the detected face keypoints are from the source image or the driving image, and $K$ represents the total number of detected keypoints. 

Based on detected facial keypoints, we can estimate a motion field between two face images. 
Specifically, we adopt the Taylor approximation method to compute the motion flow between two facial images. Then, we utilize the paired keypoints  of the two face images to estimate a group of sparse motion fields $\{\mathcal{T}_{S \leftarrow D,k}(z)\}^K_{k=1}$ for all the keypoints as follows:
\begin{equation}
    \mathcal{T}_{S \leftarrow D,k}(z) = \mathbf{x}_{S,k} + J_{S,k}J_{D,k}^{-1}(z-\mathbf{x}_{D,k}),
\end{equation}
where $z\in\mathbb{R}^2$ is an arbitrary point on the driving face. After that, we adopt the dense motion module from FOMM~\cite{siarohin2019first} to estimate a dense 2d motion field $\mathcal{T}_{S\leftarrow D}$ with the input of sparse affine motion fields $\{\mathcal{T}_{S \leftarrow D,k}(z)\}^K_{k=1}$:
\begin{equation}
    \mathcal{M}, \mathcal{T}_{S\leftarrow D} = \mathcal{F}_M \left(\{\mathcal{T}_{S \leftarrow D,k}(z)\}^K_{k=1}, \mathbf{I}_S\right),
\end{equation}
where $\mathcal{F}_M(\cdot,\cdot)$ is the dense motion module, and the estimated occlusion map $\mathcal{M}$ is used to mask out the regions of the feature map that have ambiguities due to varying and relatively large rotations of the face.

\subsection{Geometry-enhanced multi-layer generation}\label{sec:mgm}
To effectively incorporate the acquired facial depth maps for enhancing generation quality, we introduce a cross-modal (\ie~depth and image) attention mechanism, facilitating better preservation of facial structures and generation of expression-related micro facial movements. The depth information, offering dense 3D face geometry, proves to be fundamentally advantageous in maintaining facial structures and recognizing critical movements during generations. In this work, we expand DaGAN's cross-modal attention mechanism~\cite{hong2022depth} by implementing it across multiple layers, enabling the model to capture face geometric information among different appearance feature levels. 
It ultimately contributes to enhanced generation performance. As depicted in Fig.~\ref{fig:depth_attention}, each cross-modal geometry-guided attention module comprises two stages, \ie~a feature warping step and a cross-modal attention computation step. We present details of these two steps in the following.

\subsubsection{Feature warping}  
In the generation process, we consider an encode-decoder architecture to generate images. Firstly, we use an encoder to produce an $L$-layer encoded appearance feature map $\{\mathbf{F}_{e}^i\}_{i=1}^N$. Meanwhile, we employ a depth encoder $\mathcal{E}_d$ {(see Fig.~\ref{fig:structure}b)} to predict a depth feature map $\mathbf{F}_{depth}$ with the input source depth map $\mathbf{D}_s$.
The $i$-th cross-modal attention module takes the appearance feature $\mathbf{F}_{e}^i$, the depth feature $\mathbf{F}_{depth}$, the occlusion map $\mathcal{M}$, and the motion field $\mathcal{T}_{S\leftarrow D}$ as input to produce a geometry-enhanced feature map $\mathbf{F}_{en}^i$. As shown in Fig.~\ref{fig:depth_attention}a, we initially align the source view feature map $\mathbf{F}_{e}^i$ to the target view by warping $\mathbf{F}_{e}^i$ using the motion field $\mathcal{T}_{S\leftarrow D}$. This rough alignment is then followed by a multiplication of the occlusion map that indicates possible ambiguities caused by large motions as:
\begin{equation}
    \mathbf{F}_w^i = \mathcal{M}\times \mathcal{W}_p(\mathbf{F}_e^i, \mathcal{T}_{S\leftarrow D}),
\end{equation}
where $\times$ represents an element-wise multiplication operation, and $\mathcal{W}_p$ demotes the warping function. In this manner, the warped features $\mathbf{F}_w^i$ can retain the source image's identity while simultaneously preserving the head motion information between the source and target faces.

\subsubsection{Cross-attention geometry enhancement} We incorporate facial geometry information into the generation process by executing cross-modal attention between the warped feature map $\mathbf{F}_w^i$ and the depth feature map $\mathbf{F}_{depth}$. As illustrated in Fig.~\ref{fig:depth_attention}, a linear projection are performed on both $\mathbf{F}_{depth}$ and the warped source-image feature $\mathbf{F}_w^i$, resulting in $\mathbf{F}_q^i$, $\mathbf{F}_k^i$, and $\mathbf{F}_v^i$ using three distinct $1\times 1$ convolutional layers with kernels $\mathbf{W}_q^i$, $\mathbf{W}_k^i$, and $\mathbf{W}_v^i$, respectively. These maps, $\mathbf{F}_q^i$, $\mathbf{F}_k^i$, and $\mathbf{F}_v^i$, correspond to the query, key, and value in the cross-attention mechanism. 
We employ the appearance feature map $\mathbf{F}_w^i$ to produce the query $\mathbf{F}_q^i$, which shares the same shape as the output. 
As a result, geometry-related information can be queried via cross-attention from the geometry features. The geometry-related information is further added to the appearance feature map via the residual connection using an addition operation. In this way, the depth features can provide dense guidance for face generation.
In details, the geometry-enhanced features $\mathbf{F}_{en}^i$ for generation at the $i$-th layer are obtained as follows:
\begin{gather}
    \mathbf{F}_{en}^i =  \mathrm{Softmax}\left(\mathbf{F}_q^i \otimes (\mathbf{F}_k^i)^T\right) \otimes \mathbf{F}_v^i+\mathbf{F}_w^i,
\end{gather}
where $\mathrm{Softmax}(\cdot)$ is a softmax operation, and $\otimes$ indicates a matrix multiplication operation. Because of 3D geometric guidance, our model can better perceive the facial structure and micro-movements of the driving face. 

\subsubsection{Multi-layer geometry enhancement} 
To enable the model to be perceptive to the facial geometry in the whole generation process, we also apply the aforementioned cross-modal attention mechanism in each layer of the generation process. As shown in Fig.~\ref{fig:depth_attention}b, 
the final feature map $\mathbf{F}_{d}^{N}$ is generated by the concatenation of the geometry-enhanced feature $\mathbf{F}_{en}^{N}$ and the $\mathbf{F}_{d}^{N-1}$ after an upsampling block. At the $i$-th level ($i>1$), we feed the decoded feature map $\mathbf{F}_{d}^i$ into an upsample block and the upsampled result is concatenated with the geometry-enhanced feature $\mathbf{F}_{en}^{i+1}$ to produce a next-layer decoded feature $\mathbf{F}_d^{i+1}$ via a convolutional layer. 
Finally, we feed the $\mathbf{F}_d^N$ into a convolution layer, and then use a Sigmoid unit to generate the final facial image $\mathbf{I}_{rst}$. In this way, our model can realize the facial structure in each layer provided by the geometry information to produce realistic human faces.
\begin{figure*}[t]
  \centering
  \includegraphics[width=1\linewidth]{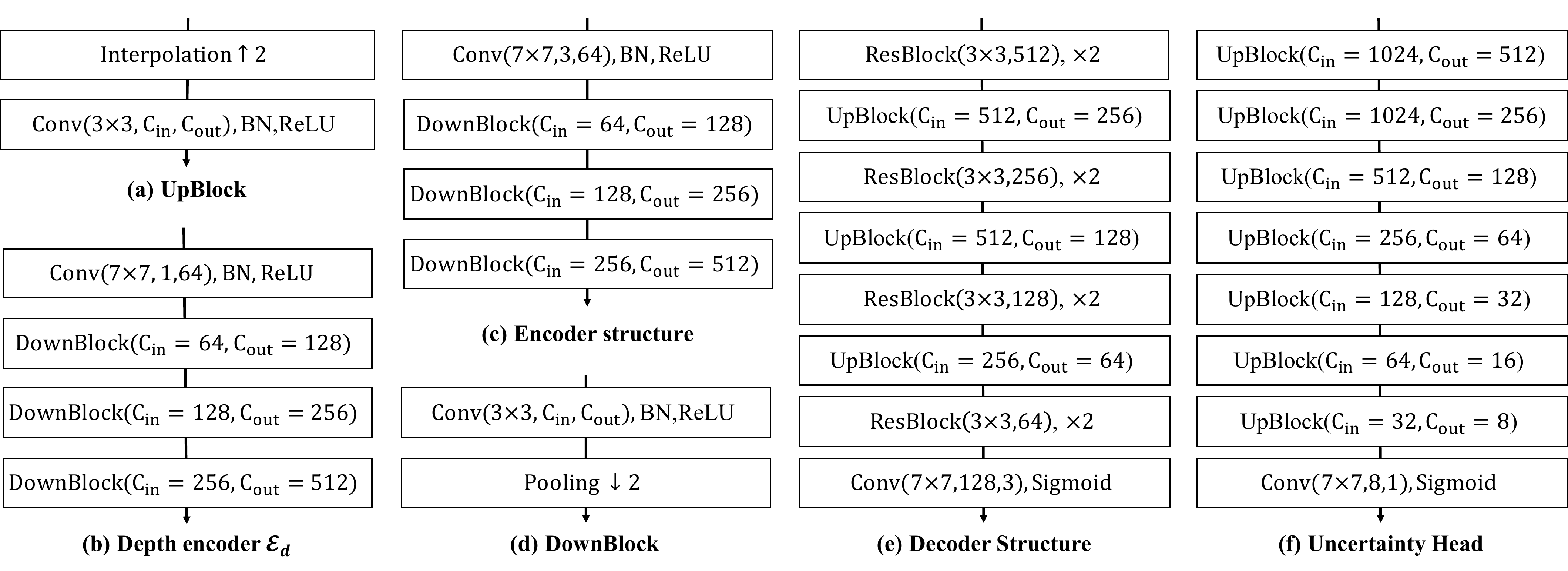}
   \caption{The detailed structures of each sub-network of DaGAN++. The ``DownBlock'' (Fig.~\ref{fig:structure}d) includes an average pooling layer and a convolutional block, which consists of a convolution layer with $3\times 3$ kernel, a batch normalization layer, and a ReLU activation layer. The interpolation layer in ``UpBlock'' (Fig.~\ref{fig:structure}a) is utilized to upsample the image. The symbol ``$\times 2$'' in other modules means that the module has been used twice.}
   \label{fig:structure}
\end{figure*}
\vspace{-5pt}
\subsection{Optimization}
In accordance with prior research~\cite{siarohin2019first, wang2021one}, we learn our DaGAN++ by minimizing the following loss:
 \begin{equation}
 \begin{aligned}
\mathcal{L} = &\lambda_P\mathcal{L}_P(\mathbf{I}_{GT},\mathbf{I}_{rst})+\lambda_E\mathcal{L}_E(\{\mathbf{I}_S||\mathbf{D}_S)\\
&+ \lambda_D(\mathcal{L}_D(\{\mathbf{x}_{S,k}\}_{k=1}^K)+\mathcal{L}_D(\{\mathbf{x}_{D,k}\}_{k=1}^K)), 
 \end{aligned}
 \end{equation}
\noindent\textbf{Perceptual loss $\mathcal{L}_P$.} We consider a pre-trained VGG-19~\cite{johnson2016perceptual} model to minimize the appearance feature difference between the ground truth face image $\mathbf{I}_{GT}$, which is the driving frame $\mathbf{I}_D$ during the training stage, and the generated face image $\mathbf{I}_{rst}$ at multi-resolutions:
\begin{equation}
    \mathcal{L}_{P} = \sum_{l,i}(|V_{i}^l(\mathbf{I}_{GT})-V_{i}^l(\mathbf{I}_{rst})|,
\end{equation}
where $V_i$ is the $i^{th}$ layer feature of the VGG-19 model, and $l$ indicates that the input is downsampled by $l$ times. 

\noindent\textbf{Equivariance loss $\mathcal{L}_E$.} Similar to FOMM~\cite{siarohin2019first}, we use the equivariance loss to guarantee the coherence of the identified keypoints.
\begin{equation}\label{eq:total_loss}
    \mathcal{L}_{E} = |\mathcal{F}_{kp}(\mathcal{T}_{random}(\mathbf{I}_\tau||\mathbf{D}_\tau)) - \mathcal{T}_{random}(\mathcal{F}_{kp}(\mathbf{I}_\tau||\mathbf{D}_\tau))|,
\end{equation}
where $\mathcal{T}_{random}$ is a random nolinear transformation. In this work, we apply the random TPS transformation similar to FOMM~\cite{siarohin2019first}. By using this loss, the keypoints detector can produce confident and coherent keypoints with any input.

\noindent\textbf{Keypoints distance loss $\mathcal{L}_D$.} We penalize the model when the distance between two keypoints is below a predetermined threshold, so that we can prevent the keypoints from clustering within a confined neighbourhood. For any pair of keypoints $\mathbf{x}_{\tau,i}$ and $\mathbf{x}_{\tau,j}$ of a face image $\mathbf{I}_\tau$, we establish the following regularization loss:
\begin{equation}
    \mathcal{L}_D =  \sum_{i=1}^K\sum_{j=1}^K (1-\mathbf{sign}(||\mathbf{x}_{\tau,i} - \mathbf{x}_{\tau,j}||_1-\beta)), i\neq j,
\end{equation}
where $\mathbf{sign}(\cdot)$ denotes a sign function, and $\beta$ represents the distance threshold. In our work, we set $\beta$ to 0.2, which shows satisfactory performance in our experiments.
The hyper-parameters $\lambda_P$, $\lambda_E$, and $\lambda_D$ facilitate balanced learning from these losses.
\vspace{-10pt}
\subsection{Network architecture details}
We introduce more details about the different network components in DaGAN++. 
We show the implementation of each component in Fig.~\ref{fig:structure} and elaborate on them below.

\noindent\textbf{Feature encoder $\mathcal{E}_I$.} As shown in Fig.~\ref{fig:structure}c, our feature encoder $\mathcal{E}_I$ consists of three downsample blocks, producing four different-scale feature maps{. Thus,} we can obtain both low-level and high-level facial feature maps, which respectively contain detailed facial textures and semantics. 

\noindent\textbf{Depth encoder $\mathcal{E}_d$.} We show the structure of the face depth encoder $\mathcal{E}_d$ as presented in Fig.~\ref{fig:structure}b. Its structure is identical to $\mathcal{E}_I$, ensuring that the features learned from both modalities possess equivalent representation power.

\noindent\textbf{Feature decoder.} In Fig.~\ref{fig:structure}e, we insert ResBlock in the middle of the generation network to boost the network's capacity. As mentioned in Section~\ref{sec:mgm}, we apply cross-modal attention as the skip connection to embed the geometry information into different layers in the generation process.

\noindent\textbf{Uncertainty Head.} In Fig.~\ref{fig:structure}f, we utilize a sequence of UpBlock to upsample the output of the pose network, which has the dimensions of $[1024\times 2\times 2]$. We utilize a sigmoid at the end of the uncertainty head to predict the uncertainty map with the values in $[0,1]$.

\section{Experiments}\label{sec:experiments}
\begin{figure*}[ht]
  \centering
  \includegraphics[width=0.98\linewidth]{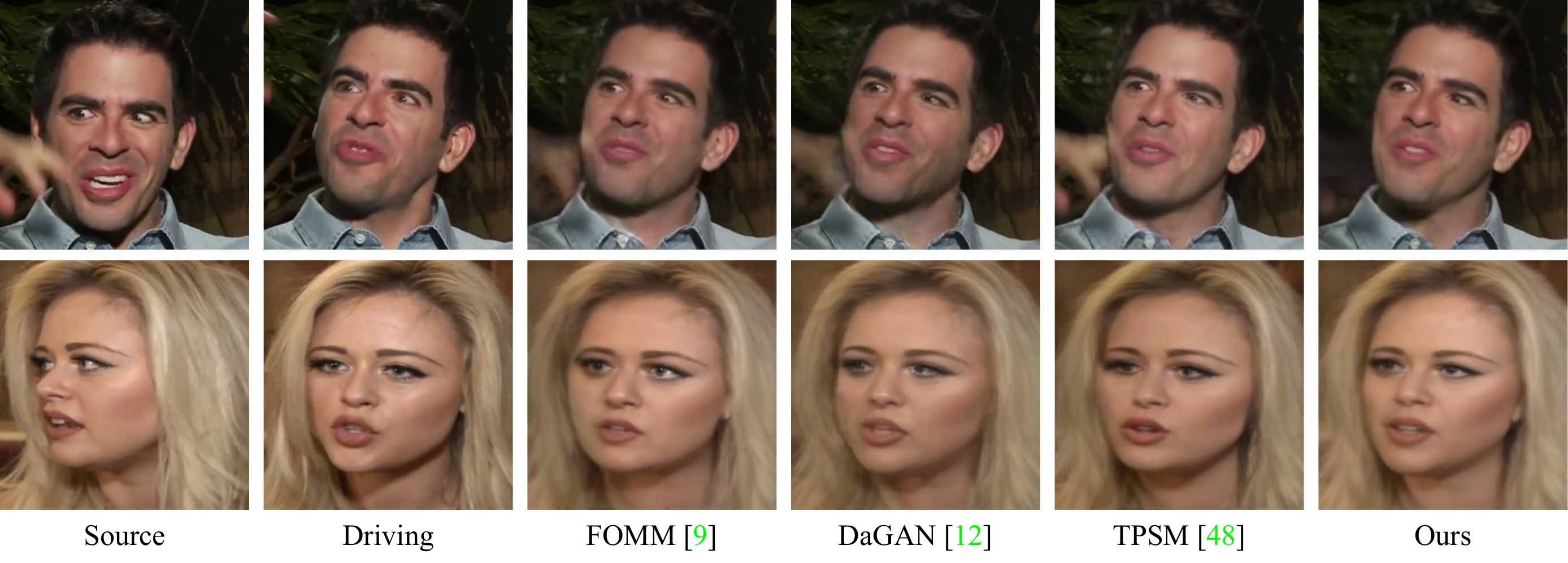}
    \vspace{-7pt}
\centering\caption{Qualitative comparisons on the self-reenactment experiment on the VoxCeleb1 dataset~\cite{nagrani2017voxceleb}. It can be observed that our method produces fewer artifacts than other competing methods (the first row) and captures better facial structures (\eg~the eye regions in the third row).}
   \label{fig:same-vox1}
   \vspace{-7pt}
\end{figure*}
We verify the effectiveness of our proposed approach through comprehensive experiments conducted on multiple publicly accessible benchmarking datasets for talking head generation. {In this section, we will illustrate the experimental setup first and then analyze our experimental results.}


%
%
\begin{figure*}[t]
  \centering
  \includegraphics[width=0.98\linewidth]{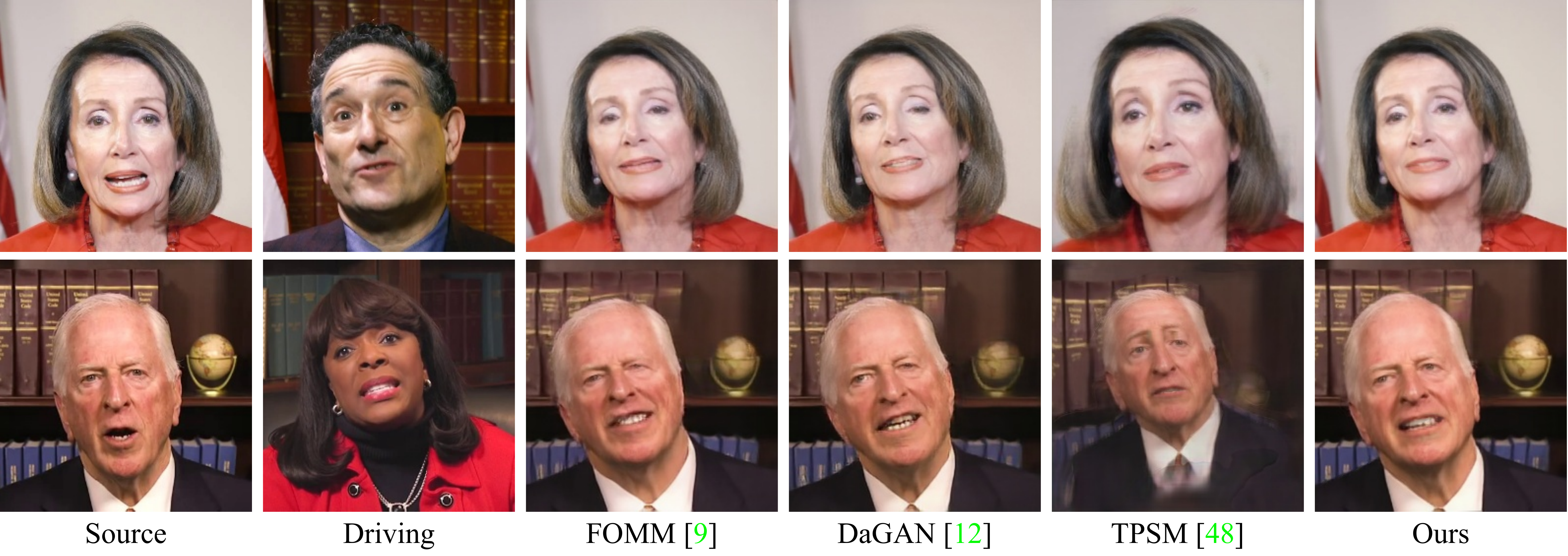}
  \vspace{-5pt}
   \centering\caption{Qualitative comparisons on the cross-identity reenactment experiments on the HDTF dataset~\cite{zhang2021flow}. DaGAN++ can capture better expression-related facial movements in the driving frame (\eg, the eye and the mouth regions), because of the enhancement by our estimated  accurate facial geometry.}
   \label{fig:cross-hdtf}
   \vspace{-10pt}
\end{figure*}
\vspace{-5pt}
\subsection{Dataset description}
We conduct experiments on three different face reenactment datasets (\ie,~ VoxCeleb1 \cite{nagrani2017voxceleb}, VoxCeleb2~\cite{chung2018voxceleb2}, and HDTF~\cite{zhang2021flow} dataset) in this work. 

\noindent\textbf{VoxCeleb1 dataset.} The VoxCeleb1 dataset is a largely collected audio-visual dataset originally for speaker identification and recognition tasks. 
The dataset includes a diverse range of speakers regarding age, gender, and ethnicity, covering a wide variety of accents and languages. It consists of a development and a test set. The development set contains $1,211$ identities, while the test set includes $40$ identities. Importantly, there is no speaker overlap between the two sets, ensuring a proper evaluation of the models on unseen data. The videos are not preprocessed for face detection or alignment, requiring models to handle the raw data. This makes the dataset challenging for talking head generation. 

\noindent\textbf{VoxCeleb2 dataset.}
VoxCeleb2~\cite{Chung18b} builds upon the original VoxCeleb1 dataset. It is designed for tasks like speaker identification, face recognition, and talking head generation. {This dataset has more than} $1$ million video clips from more than $6,000$ celebrities collected from YouTube, offering a diverse range of speakers across ages, genders, ethnicity, accents, and languages. There is no overlap between the two sets, ensuring a proper evaluation of models on unseen data. VoxCeleb2 videos are annotated with the speaker's identity and unconstrained in terms of background noise, lighting, and head poses. This provides a challenging dataset that is more representative of real-world scenarios for various computer vision and speech-processing tasks.

\noindent\textbf{HDTF dataset.} The HDTF dataset~\cite{zhang2021flow} comprises approximately 362 distinct videos, with a total of 15.8 hours of content. The original videos have resolutions of either 720P or 1080P. A landmark detector is initially employed to isolate the facial region, with the crop window remaining consistent throughout each video. Subsequently, each cropped video is resized to a dimension of $512 \times 512$ to maintain visual quality. 

\begin{figure*}[t]
  \centering
  \includegraphics[width=0.98\linewidth]{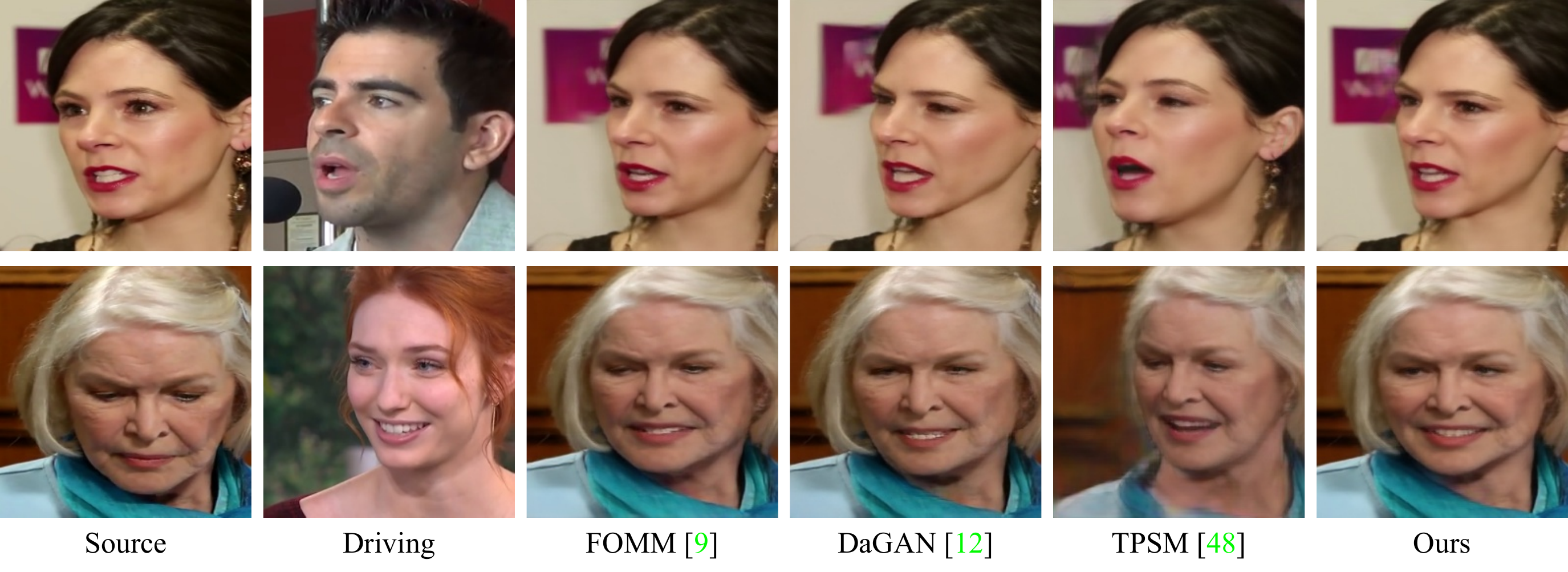}
  \vspace{-5pt}
   \centering\caption{Qualitative comparisons of the cross-identity reenactment experiment on the VoxCeleb1 dataset~\cite{nagrani2017voxceleb}. Compared with other methods, our proposed DaGAN++ can maintain the background information better (see the second row) because the depth map is beneficial for distinguishing the foreground from the background. Utilizing accurate geometry information, our method can produce natural-looking results.}
   \vspace{-5pt}
   \label{fig:cross-vox}
\end{figure*}
\vspace{-5pt}
\subsection{Metrics description}
In this research, we employ various measures to evaluate the quality of the produced images. We use structural similarity (\textbf{SSIM}), peak signal-to-noise ratio (\textbf{PSNR}), and \textbf{LPIPS} to assess the visual resemblance between the generated images and the ground-truth images. Furthermore, we incorporate three additional metrics, namely \textbf{$\mathcal{L}_1$}, Average Keypoint Distance (\textbf{AKD}), and Average Euclidean Distance (\textbf{AED}), as suggested in~\cite{siarohin2019animating}, to evaluate methods based on keypoints. \ting{Following DaGAN, we also introduce \textbf{CSIM}, \textbf{AUCON} and \textbf{PRMSE} to evaluate the qualtiy of the results under the cross-id reenactment.}

%
%
\vspace{-5pt}
\subsection{Implementation details}
During the facial depth network learning process, the architecture of the pose network and the depth network is consistent with that of DaGAN~\cite{hong2022depth}. However, we append an uncertainty prediction head at the end of the pose network to estimate the uncertainty map. The uncertainty head implementation is shown in Fig.~\ref{fig:structure}f. In this work, our keypoint detector is similar to DaGAN~\cite{hong2022depth}, while we use the same structure as that in~\cite {gordon2019depth} for implementing the face depth network and the pose networks. Regarding the optimization losses, we assign $\lambda_P$ = 10, $\lambda_E$ = 10, and $\lambda_D$ = 10. The number of keypoints in DaGAN++ is set to 15. \ting{It is noteworthy that our DaGAN++ contains 85.2M trainable parameters, while DaGAN 
 has 63.1M parameters. This increase is attributable to the geometry enhancement that we employ at each layer of the generation process, which naturally introduces additional parameters. However, when compared to state-of-the-art methods such as TPSM~\cite{zhao2022thin}, which has a similar number of trainable parameters (85.1M), our method can still obtain superior performance.
} During the training phase, we initially train our enhanced uncertainty-guided facial depth network using consecutive frames from VoxCeleb1 videos and maintain it fixed throughout the training of the entire deep generation framework.
In this work, we utilize the public code\footnote{https://github.com/AliaksandrSiarohin/video-preprocessing} to process the raw data of the Voxceleb1 dataset. 
\vspace{-5pt}
\subsection{Comparison with state-of-the-art methods}

\begin{table}[t]
\caption{Quantitative comparison of state-of-the-arts on the self-reenactment experiment on VoxCeleb1 dataset.}
  \resizebox{1\linewidth}{!}{
        \begin{tabular}{lcccccc}
        \toprule
        
         Model & SSIM (\%) $\uparrow$ & PSNR $\uparrow$  &  LPIPS $\downarrow$ & $\mathcal{L}_1$ $\downarrow$ & AKD $\downarrow$ &  AED $\downarrow$ \\
        \midrule
        X2face \cite{wiles2018x2face} &71.9 & 22.54  & - & 0.0780&7.687&0.405 \\
        NeuraHead-FF(\cite{zakharov2019few}) &63.5 & 20.82  & - & - &- & - \\
        marioNETte \cite{ha2020marionette} &75.5 & 23.24 & - & - &- & - \\
      FOMM \cite{siarohin2019first}) & 72.3 & 30.39  & 0.199  & 0.0430 & 1.294 & 0.140 \\
      MeshG \cite{yao2020mesh}& 73.9&30.39 & - & - & - & - \\
      face-vid2vid \cite{wang2021one}& 76.1& 30.69 & 0.212 & 0.0430 & 1.620 & 0.153 \\
      MRAA \cite{siarohin2021motion}& 80.0  & 31.39  & 0.195 & 0.0375 & 1.296 &0.125 \\
      TPSN \cite{zhao2022thin}& 81.6  & 31.43  & 0.179 &0.0365 & 1.233 &0.119 \\
        \midrule
        DaGAN \cite{hong2022depth} & 80.4 &31.22  & 0.185  & 0.0360 &1.279 &0.117  \\
        DaGAN++ (Ours) & \textbf{82.6} & \textbf{31.94}  & \textbf{0.175} & \textbf{0.0332} & \textbf{1.226} & \textbf{0.115} \\
        \bottomrule
        \end{tabular}}
        \vspace{-10pt}
\label{tab:same-vox1}
\end{table}

\begin{table}[t]
\caption{Quantative comparison of state-of-the-art methods on the self-reenactment experiments on the HDTF dataset. Our DaGAN++ clearly outperforms all the comparison methods including DaGAN on all the metrics.}
  \centering
  \resizebox{1\columnwidth}{!}{
        \begin{tabular}{lccccccc}
        \toprule
        Model & SSIM (\%) $\uparrow$ & PSNR $\uparrow$ &  LPIPS $\downarrow$ & $\mathcal{L}_1$ $\downarrow$ & AKD $\downarrow$ & AED $\downarrow$\\
        \midrule
        FOMM~\cite{siarohin2019first} & 76.9 & 31.87 & 0.155 & 0.0363 & 1.116 &0.092 \\
        MRAA~\cite{siarohin2021motion} & 79.4 & 32.32 & 0.156 & 0.0331 & 1.039 & 0.100 \\
        TPSM~\cite{zhao2022thin} & 86.0 & 32.85 & 0.114 & 0.0264 & 1.015 & 0.072 \\
         \midrule
         DaGAN~\cite{hong2022depth} & 82.3 & 32.29 & 0.136 & 0.0304 & 1.013 & 0.075 \\
        DaGAN++ (Ours) & \textbf{86.7} & \textbf{33.51} & \textbf{0.109}& \textbf{0.0239} & \textbf{0.968} & \textbf{0.063} \\
        \bottomrule
        \end{tabular}
}
\label{tab:same-hdtf}
\end{table}

\begin{table}[h]
\caption{Quantitative comparison of state-of-the-art methods on the self-reenactment experiment on the Voxceleb2 dataset. Our DaGAN++ outperforms all the comparison methods including DaGAN on all the metrics.}
  \centering
  \resizebox{1\linewidth}{!}{
        \begin{tabular}{lcccccc}
        \toprule
        Model & SSIM (\%) $\uparrow$ & PSNR $\uparrow$ &  LPIPS $\downarrow$ & $\mathcal{L}_1$ $\downarrow$ & AKD $\downarrow$ &  AED $\downarrow$ \\
        \midrule
        FOMM~\cite{siarohin2019first} & 77.19 & 30.71 & 0.257 & 0.0513 & 1.762 &  0.212\\
        MRAA~\cite{siarohin2021motion} & 78.07 & 30.89 & 0.262 & 0.0511 & 1.796 &  0.213 \\
        TPSM~\cite{zhao2022thin} & 78.22 & 30.63 & 0.254 & 0.0527 & 1.703 & 0.210 \\
        \midrule
        DaGAN~\cite{hong2022depth} & 79.02 & 30.81 & 0.250 & 0.0483 & 1.865 & 0.203 \\
        DaGAN++ (Ours)  & \textbf{80.15} & \textbf{31.12} & \textbf{0.244}& \textbf{0.0469} & \textbf{1.675} & \textbf{0.195} \\
        \bottomrule
        \end{tabular}
}
\vspace{-10pt}
\label{tab:same-vox2}

\end{table}
\begin{table}[h]
\caption{\ting{Quantitative comparison of state-of-the-art methods on the cross-id reenactment experiment on the Voxceleb1 dataset. Our DaGAN++ outperforms all competing methods, including DaGAN, across all the evaluation metrics. DaGAN++ also achieves the best results in terms of the Temporal Consistency Metric (TCM) in the same-reenactment experiments on the VoxCeleb1 dataset.}}
  \centering
  \resizebox{1\linewidth}{!}{
        \begin{tabular}{lccccc}
        \toprule
        Metric & FOMM~\cite{siarohin2019first} & MRAA~\cite{siarohin2021motion}& TPSM~\cite{zhao2022thin} & DaGAN~\cite{hong2022depth} & DaGAN++ (Ours) \\
        \midrule
        CSIM $\uparrow$ & 0.7330 & 0.7504 & 0.7556 &  0.7456& \textbf{0.7581}\\
        AUCON $\uparrow$ &0.7651 & 0.7130 & 0.7600 & 0.7550 & \textbf{0.7680} \\
        PRMSE $\downarrow$ & 3.0762 & 3.589 & 2.960 & 3.020& \textbf{2.949}\\
        TCM $\uparrow$ & 0.1628& 0.1632&0.1629 &0.1631 & \textbf{0.1806} \\
        \bottomrule
        \end{tabular}
}
\vspace{-10pt}
\label{tab:cross-vox}
\end{table}

\begin{figure*}[t]
  \centering
  \includegraphics[width=0.98\linewidth]{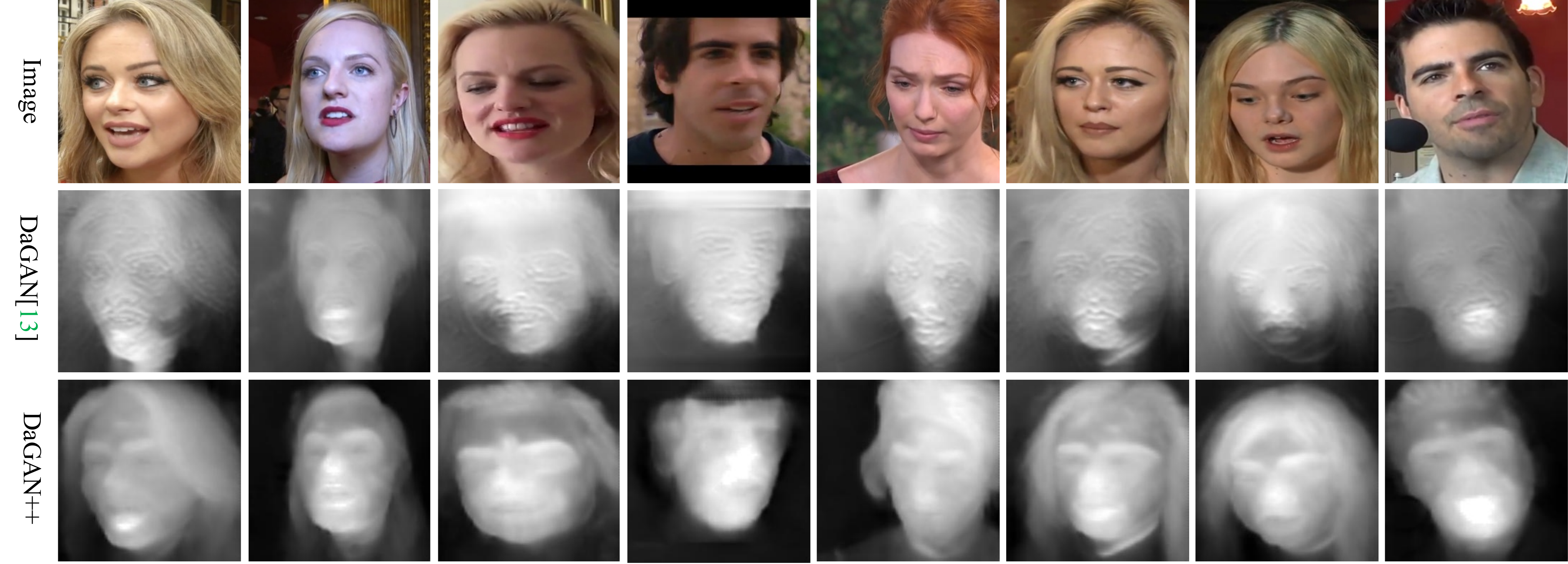}
  \vspace{-5pt}
   \centering\caption{Qualitative comparisons of depth estimation on the VoxCeleb1 dataset~\cite{nagrani2017voxceleb}. It is clear that our improved face depth network generates more accurate depth maps compared to the original depth network used in DaGAN. Our depth map can precisely estimate the background pixels and maintain clearer structural details of the foreground human faces.}
   \vspace{-5pt}
   \label{fig:depthvox1}
\end{figure*}
\begin{figure*}[t]
  \centering
  \includegraphics[width=0.98\linewidth]{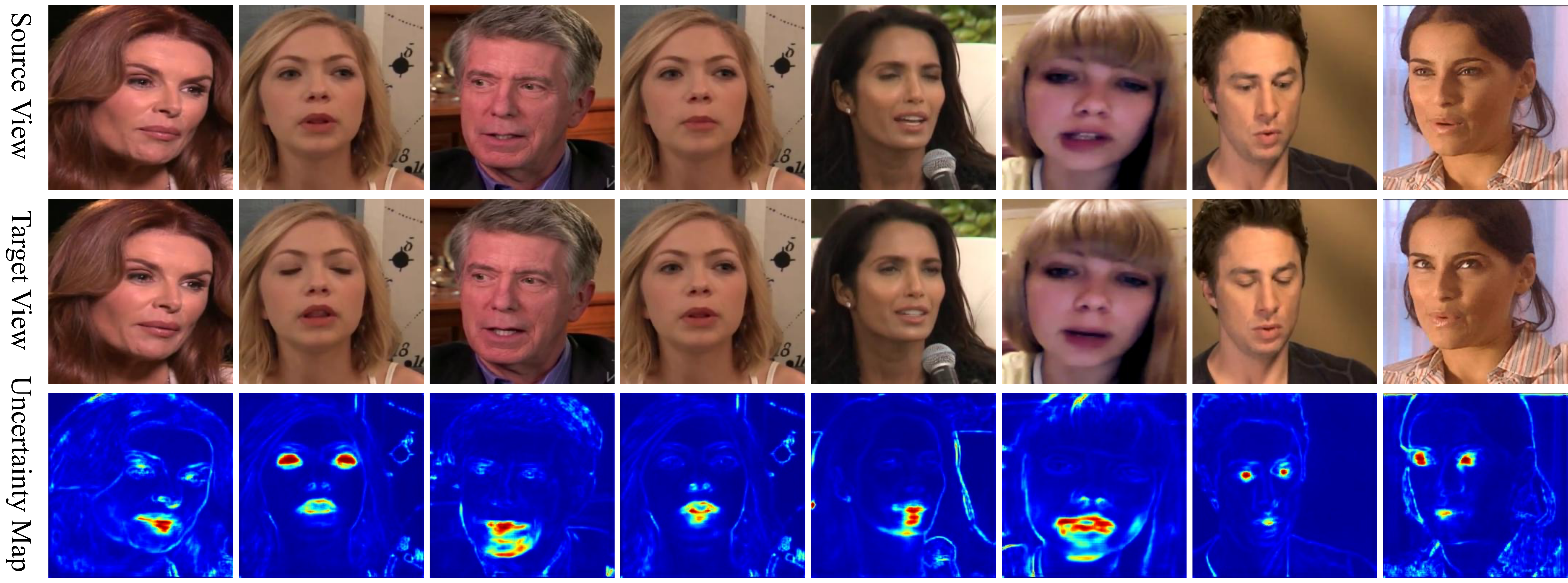}
  \vspace{-5pt}
   \caption{Examples of the learned uncertainty maps for facial depth network learning. {The pixel value of the uncertainty map assigns lower uncertainty values to the rigid part of the face-moving pixels, while assigning higher values to the non-rigid region. By incorporating the uncertainty map, our model can filter out the non-rigid-motion pixels to better learn the dominant rigid motion between two views for depth learning.}}
   \label{fig:uncertainty_map}
   \vspace{-5pt}
\end{figure*}

\noindent\textbf{Self-reenactment.} Firstly, we conduct the self-reenactment experiment to quantitatively evaluate our method. The results of the self-reenactment are reported in Table~\ref{tab:same-vox1}, Table~\ref{tab:same-vox2}, and Table~\ref{tab:same-hdtf}. We can observe that DaGAN++ gains the best performance compared with other methods on all the datasets. 
Compared with the DaGAN, our DaGAN++ gains a significant improvement in most of the metrics, \eg, $86.7\%$ vs $82.3\%$ on the SSIM on the HDTF dataset. It verifies the effectiveness of our improved designs upon DaGAN. 
In comparison to other model-free approaches (FOMM~\cite{siarohin2019first}, MRAA~\cite{siarohin2021motion}, and TPSM~\cite{zhao2022thin}), our DaGAN++ model is capable of capturing precise head movement. Our method achieves a score of $1.675$ on the AKD metric (lower is better) on the Voxceleb2 dataset, while the best score achieved by the comparison methods is $1.703$. This outcome confirms that our geometry-guided facial keypoints estimation can be used to more effectively generate the motion field between human faces. Moreover, our method continues to attain the best score of $82.6\%$ in terms of SSIM on Voxceleb1. This demonstrates that the proposal can recover finer facial details, such as expression and motion. Additionally, our method achieves top scores in other metrics, namely LPIPS, $\mathcal{L}_1$ error, and PSNR, indicating that our approach can generate more lifelike images compared to those competing methods. Beyond the quantitative comparison, we also visualize some samples in Fig.~\ref{fig:same-vox1}. From the generated faces in Fig.~\ref{fig:same-vox1}, we can observe that our method produces more accurate results. Besides, our approach generates fewer artifacts, as can be seen in the first row in Fig.~\ref{fig:same-vox1}. The is mainly because of the robust facial depth information, which can help to learn better motion representations, suppress the noise from the background, and preserve 3D facial structures.
 Overall, the superior performance of DaGAN++ consistently shown across different datasets highlights its effectiveness in handling same-identity reenactment tasks compared to the competing methods. 

\noindent\textbf{Cross-identity reenactment.} We also conduct experiments to investigate cross-identity motion transfer, where the source and driving images originate from different individuals. We qualitatively compare our method with others and show the visualizations in Fig.~\ref{fig:cross-hdtf} and Fig.~\ref{fig:cross-vox}.
For the faces in Fig.~\ref{fig:cross-vox} and Fig.~\ref{fig:cross-hdtf}, our approach generates face images with finer details compared to the other methods, such as the eye regions in the three rows in Fig.~\ref{fig:cross-hdtf}. This confirms that employing depth maps allows the model to recognize subtle facial expression movements. Our method can also produce visually appealing results for previously unseen targets. Notably, TPSM generates inferior results because it does not incorporate relative motion transfer~\cite{siarohin2019first} in facial animation. In comparison to DaGAN, the enhanced DaGAN++ produces clearly fewer artifacts while preserving the original background information from the source image, as shown in the second row in Fig.~\ref{fig:cross-vox}.
\ting{Additionally, we also report the quantitative results in Table~\ref{tab:cross-vox} and our method outperforms all the comparison methods on all the metrics.}
\begin{figure*}[t]
  \centering
  \includegraphics[width=0.98\linewidth]{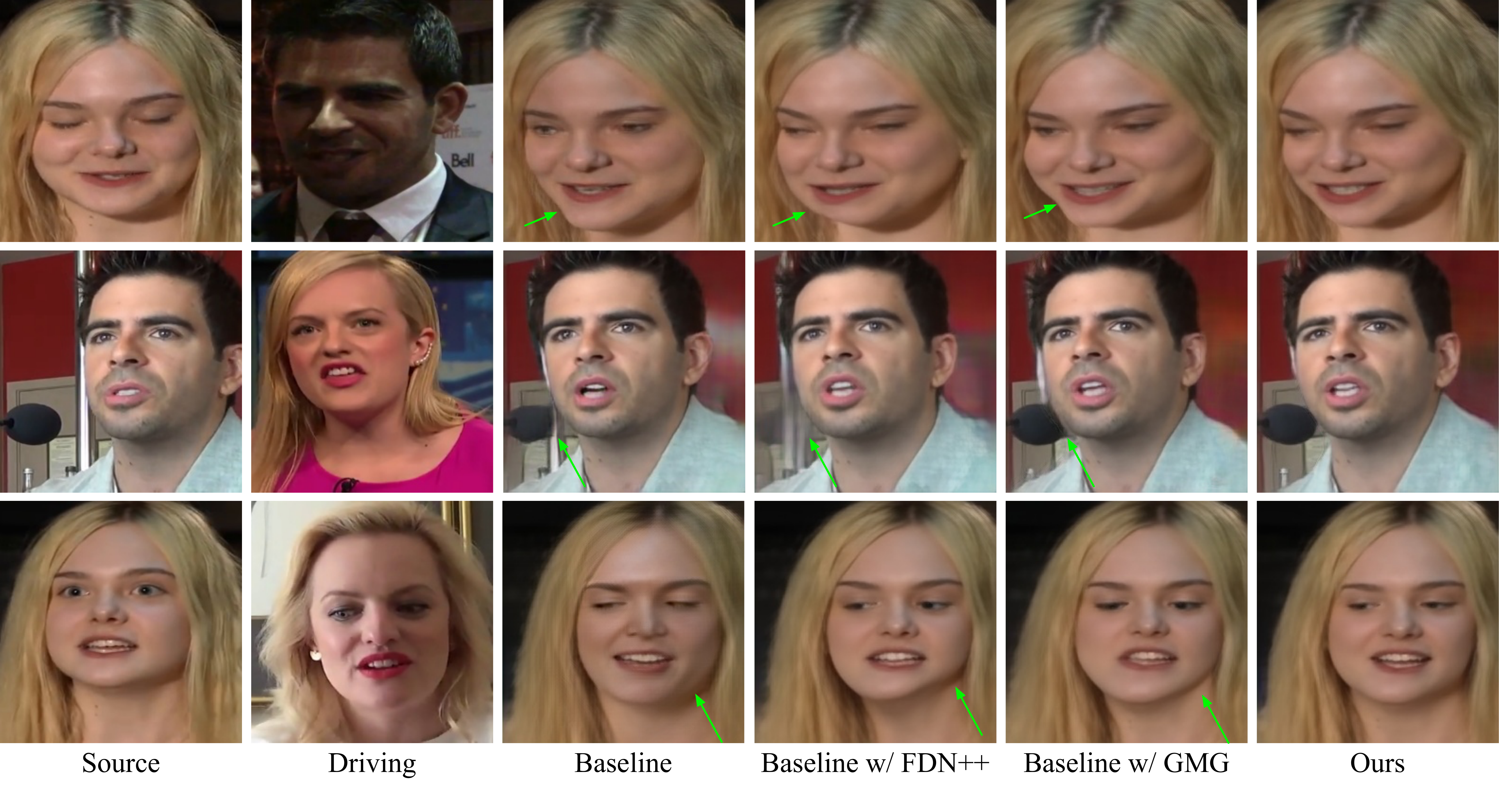}
  \vspace{-10pt}
   \centering\caption{Qualitative comparisons of different ablation studies. We can observe that DaGAN++ can gradually improve the generation results by adding the contribution modules. The final full model achieves the highest-fidelity generations.}
   \label{fig:abla}
   \vspace{-10pt}
\end{figure*}
        

  

\vspace{-5pt}
\subsection{Ablation study}
\begin{figure}[t]
  \centering
  \includegraphics[width=0.98\linewidth]{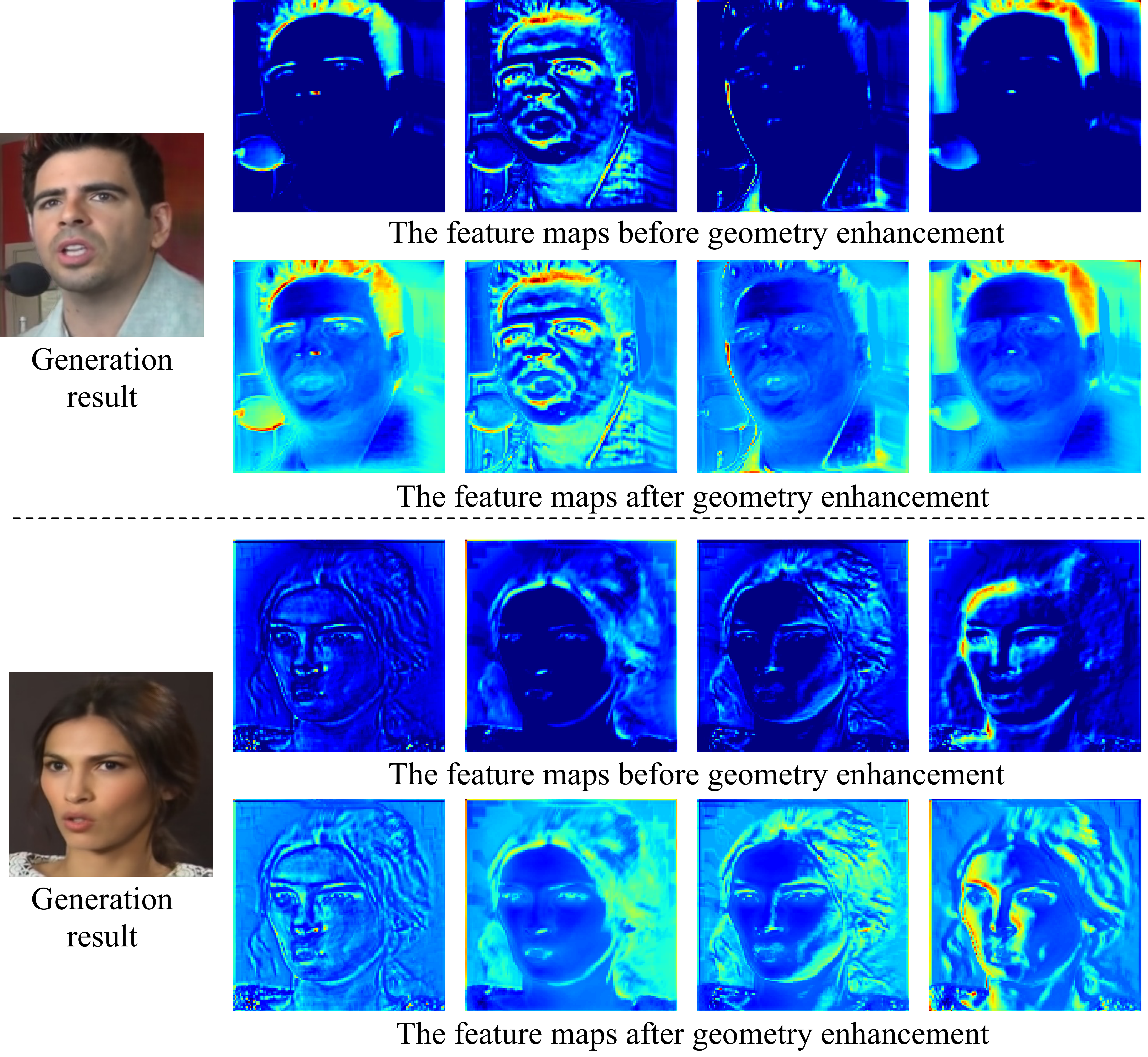}

   \caption{Investigation on the intermediate feature maps during generation. The features after our geometry enhancement can preserve more facial structures and details, and better perceive the background regions of the scene.}
   \label{fig:feature-compare}
   \vspace{-5pt}
\end{figure}

{We also perform ablation studies to validate each component of our work. \ting{In this section, we first evaluate the quality of our generated depth maps and then investigate the effect of the learned depth maps on the talking head generation.}}
We show the results in Fig.~\ref{fig:depthvox1}, Fig.~\ref{fig:uncertainty_map}, Fig.~\ref{fig:abla}, Fig.~\ref{fig:feature-compare}, Fig.~\ref{fig:out-of-domain}, Table~\ref{tab:abla} and Table~\ref{tab:hyper-kp}. In Table~\ref{tab:abla}, ``\textbf{w/~GMG}'' refers to the utilization of the multi-layer cross-modal attention mechanism to incorporate geometry information into the generation process; ``\textbf{w/~FDN++}'' denotes the application of the improved face depth network for predicting the face depth map for geometry-guided sparse keypoints estimation. In this context, our baseline corresponds to the basic model that is trained without the face depth maps and the cross-modal attention module.
\subsubsection{\ting{Self-supervised facial depth learning}}
\noindent\textbf{Depth results.}
In this study, we present improved designs for learning a more accurate facial depth estimation network compared to DaGAN. Due to the absence of ground truth data, we employ a self-supervised method for facial depth estimation. It is tricky to quantitatively evaluate the depth estimation performance due to the unavailability of ground-truth depth data, and thus the depth map's quality can only be assessed qualitatively through visualization. We show the estimated depth maps of DaGAN and DaGAN++ in Fig.~\ref{fig:depthvox1}. It is obvious that DaGAN++ produce more facial structure details in the depth maps compared to DaGAN. Furthermore, DaGAN++'s depth maps successfully eliminate the background, as demonstrated in the fifth column of Fig.~\ref{fig:depthvox1}. These findings suggest that our self-supervised facial depth learning technique can effectively recover the dense 3D geometry of human faces, which can be advantageous for subsequent talking head generation tasks. 

\noindent\textbf{Uncertainty map learning.} In our self-supervised depth learning method, we propose to learn an uncertainty map to indicate the probability of reliable and dominated {rigid-motion pixel} region of the face image. We report several samples in Fig.~\ref{fig:uncertainty_map}. Recalling the form of the depth learning objective (see Equation~\ref{eq:depth_loss}), it will try to set uncertainty $\mathcal{U}$ to regularize the photometric consistency error. As shown in Fig.~\ref{fig:uncertainty_map}, we can observe that pixels that occur more reliably {rigid} motions will be learned to assign smaller uncertainties, and then produce larger weights for the photometric consistency errors of those pixels. It verifies our motivation by learning the uncertainty map to represent {rigid and non-rigid} moving pixels with different important scores. 

\begin{table}[t]
  \centering
  \caption{Ablation studies on the HDTF dataset. `GMG' indicates the utilization of the proposed cross-modal attention module for a multi-layer geometry-enhanced generation; `FDN' and `FDN++' denote the original depth network in DaGAN and the improved one in DaGAN++, respectively; {`CAM' indicates the cross-attention module we only apply on the lowest-level feature map as done in DaGAN, while the `CAM++' is an enhanced version upon `CAM' which uses more accurate face depth map produced by our improved face depth network as input.}
  `Baseline' denotes the simplest model trained without the face depth network and cross-modal attention module in each layer. 
  }
  \vspace{-5pt}
  \resizebox{1\columnwidth}{!}{
        \begin{tabular}{lcccc}
        \toprule
        Model  & SSIM (\%) $\uparrow$ & PSNR $\uparrow$ &  LPIPS $\downarrow$ & $\mathcal{L}_1$ $\downarrow$ \\
        \midrule
        Baseline  & 81.0 & 31.30& 0.145& 0.0317\\
        Baseline w/ FDN & 81.3 & 31.57& 0.144&0.0308 \\
        Baseline w/ CAM &82.0 & 32.13 & 0.139 & 0.0307\\
        DaGAN \cite{hong2022depth} & 82.3 & 32.29 & 0.136 & 0.0304 \\
        \midrule
        \midrule
        
        Baseline w/ FDN++ & 84.3 & 32.89& 0.131 & 0.0269 \\
        Baseline w/ CAM++ & 83.2 & 32.71& 0.140 & 0.0275 \\
        Baseline w/ GMG & 86.0 & 33.21 & 0.110 & 0.0241 \\
        DaGAN++ (Ours)  & \textbf{86.7} & \textbf{33.51} & \textbf{0.109}& \textbf{0.0239} \\
        \bottomrule
        \end{tabular}
}
  
\label{tab:abla}
\vspace{-10pt}
\end{table}




\noindent\textbf{Effectiveness of improved depth maps.} As previously mentioned, accurate facial geometry information can greatly facilitate the preservation of facial structures. In this paper, we propose an enhanced face-depth learning method that substantially improves the performance compared to DaGAN, as illustrated in Fig.~\ref{fig:depth_estimate}. To quantitatively verify the effectiveness of our improved face depth network, we replace the original face depth network with the enhanced one in DaGAN. As demonstrated in Table~\ref{tab:abla}, {the component of DaGAN (\ie,~``FDN'' and ``CAM'') with our enhanced face depth network, \ie,~``FDN++'' and ``CAM++'', achieves a noticeable improvement compared to the original DaGAN, confirming the effectiveness of the enhanced facial depth estimation network. These results indicate that accurate facial geometry can indeed improve the quality of talking head generation.}

\noindent\textbf{Effectiveness of depth-guided keypionts estimation.} 
As discussed in the main paper, face depth is critically important for the keypoint estimation.
We first quantitatively investigate the influence of depth maps on keypoint detection and present the corresponding results in Table~\ref{tab:abla}. From Table~\ref{tab:abla}, it is evident that the depth-guided keypoints stably contribute to our model's performance across all the evaluation metrics compared to the baseline model, confirming the importance of depth maps for face keypoint estimation. From Fig.~\ref{fig:abla}, the "Baseline w/ FDN++" also shows a more accurate generation of facial expressions compared to the "Baseline", which suggests that the geometry-guided facial keypoints can help to produce more precise motion fields of human faces effectively. 
 
\subsubsection{\ting{Geometry-enhanced multi-layer generation}}
\noindent\textbf{Effectiveness of geometry-enhanced multi-layer module .} 
By comparing the performances between the ``baseline'' and ``Baseline w/ CAM'', we can verify the effectiveness of the cross-modal attention mechanism for talking head generation. 
In this work, we further deploy the cross-modal attention unit in each layer of the generation process, so that we can introduce geometry information into different levels of features for coarse-to-fine geometry-guided generation. From Table~\ref{tab:abla} and Fig.~\ref{fig:abla}, the geometry-enhanced multi-layer generation (GMG) can clearly {boost the quality of the output human faces}. 
Moreover, benefiting from the face geometry, our method can preserve the expression-related micro-motion of the face. 
Compared to the ``Baseline'', the ``Baseline w/ GMG'' can produce more realistic-looking expressions (\eg~at eye regions).
From Table~\ref{tab:abla} and Fig.~\ref{fig:abla}, we can verify that our proposed geometry-enhanced multi-layer generation process can effectively embed the face geometry into the generation process to capture critical movement (\eg, expressions) of the human face, thus leading to higher-quality face generation.

\noindent\textbf{Impact of depth maps on generation features.} We also visualize the intermediate generation feature maps from the generator to verify the impact of our geometry enhancement. We show the generation features before and after applying the geometry enhancement module. \ting{As mentioned in Section~\ref{sec:mgm}, the geometry enhancement enables the intermediate feature map to preserve global facial structures and recognize expression-related micro facial movements.}  \ting{By leveraging dense depth information, we can provide dense (pixel-wise) geometry enhancement across the entire feature map. This improves the global facial structure rather than employing region-focused attention that highlights only specific local areas. Therefore, the feature map with geometry enhancement offers more texture details across the entire image as shown in Fig.~\ref{fig:feature-compare}.}

\begin{figure}[t]
  \centering
    \includegraphics[width=1\linewidth]{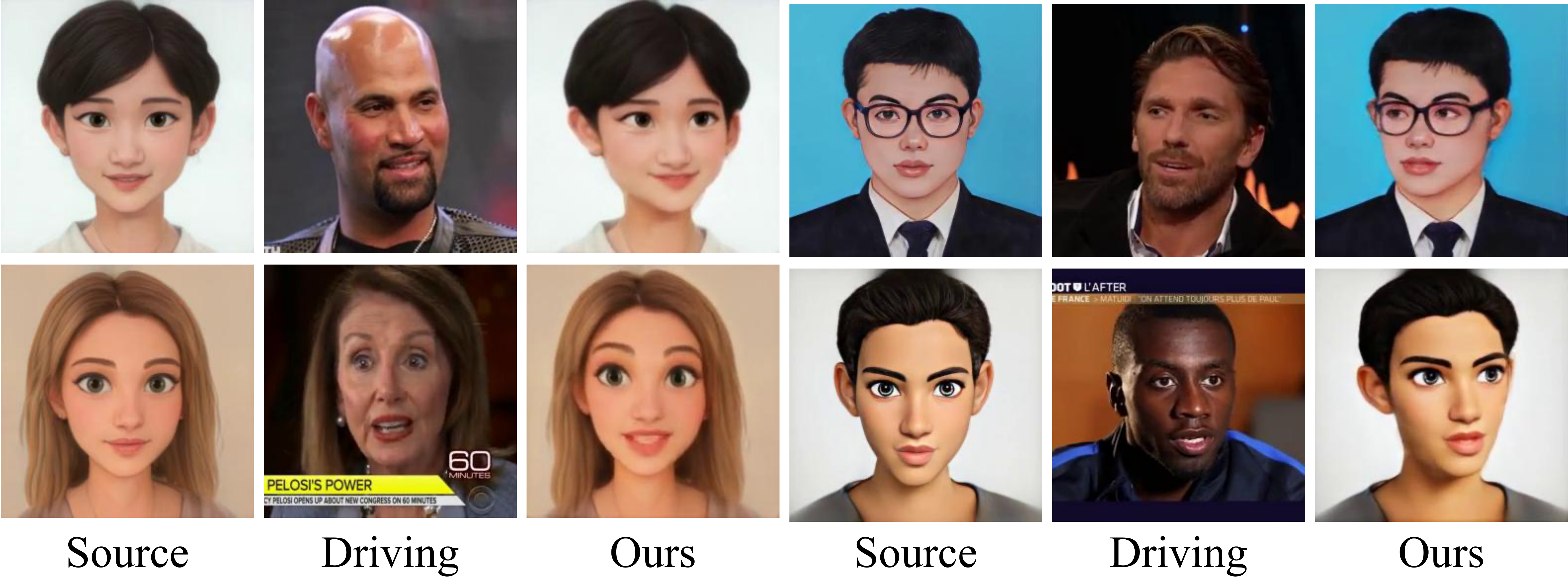}
    \caption{Qualitative results of out-of-domain generation.}
    \label{fig:out-of-domain}     
\end{figure}

\noindent\textbf{Out-of-domain results.} 
\ting{In addition to real human faces, we present our generation results for some out-of-domain samples in Fig. \ref{fig:out-of-domain} to validate the out-of-domain generation capabilities of our model. As depicted in Fig.~\ref{fig:out-of-domain}, our method can effectively modify the expressions of faces in oil paintings or cartoons. }

\begin{table}[t]
  \centering
  \caption{\ting{Ablation studies of different keypoint numbers on the HDTF dataset.}}
  \vspace{-5pt}
  \resizebox{1\columnwidth}{!}{
        \begin{tabular}{lcccc}
        \toprule
        Model  & SSIM (\%) $\uparrow$ & PSNR $\uparrow$ &  LPIPS $\downarrow$ & $\mathcal{L}_1$ $\downarrow$ \\
        \midrule
        kp=10  & 86.3 & 33.40& 0.111& 0.0245\\
        kp=20  & 86.6 & 33.50& 0.109& 0.0240\\
        kp=15 (Ours)  & \textbf{86.7} & \textbf{33.51} & \textbf{0.109}& \textbf{0.0239} \\
        
        \bottomrule
        \end{tabular}
}
  
\label{tab:hyper-kp}
\vspace{-10pt}
\end{table}
\noindent\textbf{The selection of keypoint number.} \ting{From Table~\ref{tab:hyper-kp}, we set the number of keypoints as 15 because it can achieve the best performance. Using fewer than 15 keypoints results in sparser facial motion representation and decreased performance, while increasing the number of keypoints beyond 15 does not yield a significant performance improvement yet leads to a larger model size. One possible reason is that, as the number of keypoints increases, their collective constraints can lead to more unstable motion in specific regions.}

\noindent\textbf{Evaluation of temporal consistency.} \ting{Moreover, we also adopt a temporal consistency metric to measure and evaluate the frame-wise consistency (\ie, TCM~\cite{varghese2020unsupervised}) of the generated videos. Specifically, the TCM is defined as:}
\begin{equation}
    TCM = \frac{1}{T}\sum_{t=1}^{T}\exp{-\frac{||\mathbf{I}_{GT}^t-\mathcal{W}_p(\mathbf{I}_{GT}^{t-1},\mathcal{T}_{t-1\leftarrow t})||}{||\mathbf{I}_{rst}^t-\mathcal{W}_p(\mathbf{I}_{rst}^{t-1},\mathcal{T}_{t-1\leftarrow t})||}-1}
\end{equation}
\ting{Where \( \mathbf{I}_{\text{GT}}^t \) and \( \mathbf{I}_{\text{rst}}^t \) are the \( t^{\text{th}} \) frame of the ground-truth and the generated video, respectively. \( \mathcal{W}_p(\cdot,\cdot) \) represents the warping function, and \( \mathcal{T}_{t-1 \leftarrow t} \) is the motion flow between \( \mathbf{I}_{\text{GT}}^t \) and \( \mathbf{I}_{\text{GT}}^{t-1} \). We employ Gunnar Farneback's algorithm in OpenCV~\cite{bradski2000opencv} to calculate the motion \( \mathcal{T}_{t-1 \leftarrow t} \). Through this equation, the generated video is encouraged to maintain temporal consistency with variations in the ground-truth video. The results are shown in Table~\ref{tab:cross-vox}. Our method achieves the best TCM value because the consecutive frames have almost the same depth map estimated by our face depth network and the depth map is sufficiently stable to constrain the facial structure during the generation, thereby enabling better temporal consistency.}

\section{Conclusion}\label{sec:conclusion}
In this work, we proposed a DaGAN++ deep framework for talking head generation, which extended the depth-aware generative adversarial network (DaGAN) by largely improving the face depth estimation network and applying a multi-layer cross-modal attention mechanism in the generation process. First, we introduced an uncertainty head to predict more reliable and dominant {rigid-motion pixel regions}, which can be utilized to more effectively learn the geometry. Then, we proposed a cross-modal (appearance and depth) attention mechanism and plugged it into each layer of the generation process, to have geometry-guided coarse-to-fine generation. We evaluated our DaGAN++ on three different public benchmarks, and our proposal demonstrated superior performances compared to state-of-the-art counterparts on all the challenging datasets.

\section*{Acknowledgement}
This research is supported in part by HKUST-Zeekr joint
research funding, the Early Career Scheme of the Research
Grants Council (RGC) of the Hong Kong SAR under grant
No. 26202321, and HKUST Startup Fund No. R9253.


%



\ifCLASSOPTIONcompsoc

\ifCLASSOPTIONcaptionsoff
  \newpage
\fi



%
\bibliographystyle{ieeetr}
\bibliography{tpami}


%
\begin{IEEEbiography}[{\includegraphics[width=1in,height=1.25in,clip,keepaspectratio]{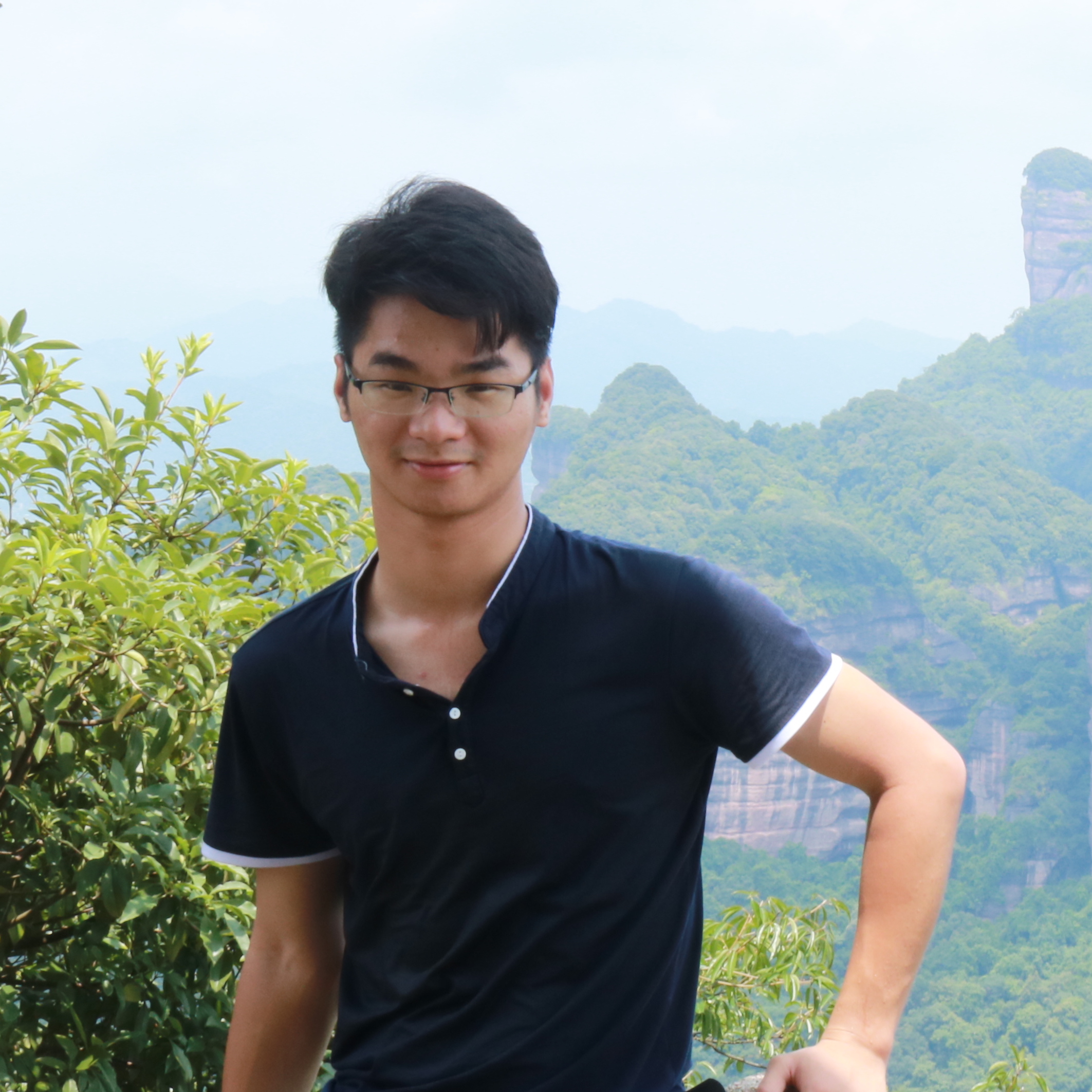}}]{Fa-Ting Hong} is a Ph.D. student in the Department of Computer Science and Engineering at Hong Kong University of Science and Technology. He received his M.E. from Sun Yat-sen University and B.E. from South China University of Technology. His research interest lies in deep learning for image and video generation.
\end{IEEEbiography}
\vspace{-10pt}
\begin{IEEEbiography}[{\includegraphics[width=1in,height=1.25in,clip,keepaspectratio]{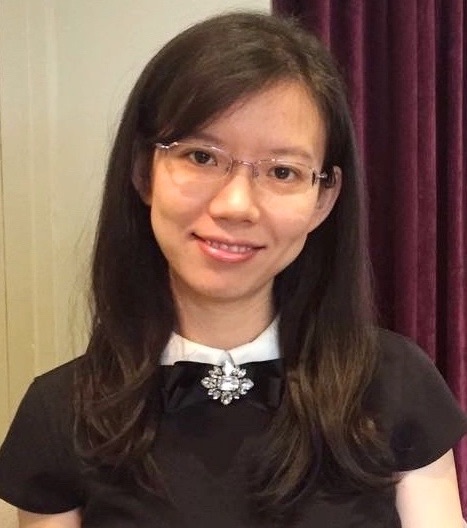}}]{Li Shen} is Staff Researcher at Alibaba Group, where she works on 3D-aware representation learning for 3D scene reconstruction and synthesis. She received her PhD in Computer Science from the University of Chinese Academy in 2016. Her research interests include computer vision and representation learning, with a focus on image recognition and generation and neural network architectures.
\end{IEEEbiography}
\vspace{-10pt}
\begin{IEEEbiography}[{\includegraphics[width=1in,height=1.25in,clip,keepaspectratio]{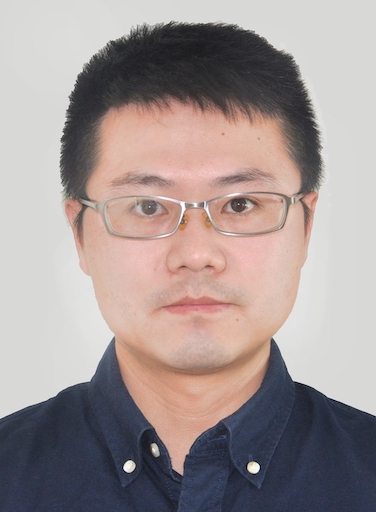}}]{Dan Xu} is an Assistant Professor in the Department of Computer Science and Engineering at HKUST. He was a Postdoctoral Research Fellow in VGG at the University of Oxford. He was a Ph.D. in the Department of Computer Science at
the University of Trento. He was also a student research
assistant at MM Lab at the Chinese University of
Hong Kong. He received the best scientific paper
award at ICPR 2016, and a Best Paper Nominee
at ACM MM 2018. He served as Area Chair at multiple main-stream conferences including CVPR, AAAI, ACM Multimedia, WACV, ACCV, and ICPR.
\end{IEEEbiography}







\end{document}